\documentclass[nohyperref]{article}

\usepackage{microtype}
\usepackage{graphicx}
\usepackage{booktabs} 
\usepackage{bm}
\usepackage{hyperref}

\usepackage[accepted]{icml2022}

\usepackage{subfiles}
\usepackage{amsmath}
\usepackage{amssymb}
\usepackage{mathtools}
\usepackage{amsthm}
\usepackage{gensymb}
\usepackage{dsfont}
\usepackage{caption}
\usepackage{subcaption}
\usepackage[capitalize,noabbrev]{cleveref}

\theoremstyle{plain}
\newtheorem{theorem}{Theorem}[section]

\theoremstyle{definition}

\theoremstyle{remark}

\usepackage[textsize=tiny]{todonotes}

\icmltitlerunning{How Tempering Fixes Data Augmentation in Bayesian Neural Networks}

\begin{document}

\twocolumn[
\icmltitle{How Tempering Fixes Data Augmentation in Bayesian Neural Networks}



\icmlsetsymbol{equal}{*}

\begin{icmlauthorlist}
\icmlauthor{Gregor Bachmann}{equal,yyy}
\icmlauthor{Lorenzo Noci}{equal,yyy}
\icmlauthor{Thomas Hofmann}{yyy}
\end{icmlauthorlist}

\icmlaffiliation{yyy}{Department of Computer Science, ETH Z\"urich, Z\"urich, Switzerland}

\icmlcorrespondingauthor{Gregor Bachmann}{gregor.bachmann@inf.ethz.ch}
\icmlcorrespondingauthor{Lorenzo Noci}{lorenzo.noci@inf.ethz.ch}

\icmlkeywords{Machine Learning, ICML}

\vskip 0.3in
]



\printAffiliationsAndNotice{\icmlEqualContribution} 

\begin{abstract}
While Bayesian neural networks (BNNs) provide a sound and principled alternative to standard neural networks, an artificial sharpening of the posterior usually needs to be applied to reach comparable performance. This is in stark contrast to theory, dictating that given an adequate prior and a well-specified model, the untempered Bayesian posterior should achieve optimal performance. Despite the community's extensive efforts, the observed gains in performance still remain disputed with several plausible causes pointing at its origin. While data augmentation has been empirically recognized as one of the main drivers of this effect, a theoretical account of its role, on the other hand, is largely missing. In this work we identify two interlaced factors concurrently influencing the strength of the cold posterior effect, namely the correlated nature of augmentations and the degree of invariance of the employed model to such transformations. By theoretically analyzing simplified settings, we prove that tempering implicitly reduces the misspecification arising from modeling augmentations as i.i.d. data. The temperature mimics the role of the effective sample size, reflecting the gain in information provided by the augmentations. We corroborate our theoretical findings with extensive empirical evaluations, scaling to realistic BNNs. By relying on the framework of group convolutions, we experiment with models of varying inherent degree of invariance, confirming its hypothesized relationship with the optimal temperature.
\end{abstract}

\section{Introduction}
\label{sec:intro}
Deep learning has lead to tremendous advances in a variety of tasks such as computer vision \citep{resnet2016}, natural language processing \citep{devlin2019bert} and reinforcement learning \citep{alphago} to name but a few. While such deep models exhibit astonishing predictive power, their black-box nature renders uncertainty estimation very difficult and often leads to over-confident decisions \citep{Nguyen2015DeepNN, szegedy2013intriguing}. To overcome this difficulty, Bayesian neural networks (BNNs) have been introduced, combining the functional form of deep models with the framework of Bayesian inference \citep{graves2011practicalvb, blundell2015weightuncertainty, hernandez2015probabilisticbp}. By forming a posterior distribution over the model parameters instead of a point estimate, uncertainty estimations become significantly better calibrated, leading to more informed decisions in safety-critical applications. Moreover, due to the multi-modal nature of neural loss landscapes \citep{NEURIPS2018_be3087e7, landscape}, Bayesian models are particularly well-suited as they naturally form an ensemble of models \citep{wilson2020case}.\\[2mm] Recently however, \citet{wenzel2020good} observed that BNNs evaluated on standard benchmarks yield suboptimal performance, even being outperformed significantly by a simple SGD baseline. They noticed on the other hand that an artificial sharpening of the posterior distribution (so-called \textit{cold} posteriors) leads to very strong performance. The sub-optimality of Bayesian models is worrisome as the Bayesian framework is equipped with theoretical guarantees regarding its optimality \citep{kolmogorov1960foundations, Savage1954, jaynes_2003}. As a consequence, a multitude of works have been put forth, exploring different potential causes of this so-called cold posterior effect (CPE), including the curated nature of standard datasets \citep{aitchison2020statistical}, the non-Bayesian nature of data augmentation \citep{izmailov2021bayesian} as well as possibly poorly chosen priors \citep{noci2021disentangling, fortuin2021bayesian}. Data augmentation has been observed to play a particularly pronounced role, being instrumental in causing the cold posterior effect in a variety of settings \citep{izmailov2021bayesian, noci2021disentangling}. \citet{nabarro2021data} develop a formalism to incorporate data augmentation into the model but unfortunately the CPE still persists. In this work we aim to bridge this gap by mathematically analyzing the effect of data augmentation on the resulting posterior. Inspired by the seminal works on ``longitudinal" data \citep{longitud_cor, Laird1982RandomeffectsMF, longitud_cor_2}, we approach the process of data augmentation in a similar spirit, taking into account the strong statistical dependence between augmented examples which breaks the i.i.d. assumption implicit in most Bayesian inference pipelines. By incorporating the correlation structure into the model, we prove that tempering approximates the correct Bayesian posterior and even matches it in simplified settings. Intuitively, the temperature plays the role of the effective sample size, adjusting for the fact that data augmentation leads to a sample size that lies between the original and the augmented one. 

We perform exhaustive experiments to validate our theoretical insights. In particular, by relying on the framework of group convolutions  \citep{cohen2016group}, we design architectures for which the invariance with respect to certain augmentations is approximately built into the model. In turn, we observe a clear correlation between the degree of model invariance and optimal temperature. Under the view of tempering as adjusting the effective sample size of the augmented dataset, using approximately invariant models decreases the need for cold posteriors, as the effective sample size is close to that of the original dataset.

In summary, our contributions are the following:
\begin{itemize}
    \item We identify the source of misspecification introduced by data augmentation in a Bayesian context. We show how augmenting samples, in conjunction with the model's invariance, induces highly correlated errors, rendering the implicit i.i.d. assumption underlying standard BNN pipelines incorrect. 
    \item We show in simplified settings that tempering the (wrong) posterior can alleviate the misspecification, resulting in a potential explanation of the CPE. 
    \item We test our theory using group convolutions, confirming our hypothesis and theoretical results on the role of invariance. Furthermore, we show that BNNs with group convolutions outperform the standard convolution counterpart in the considered settings. 
    \end{itemize}

\section{Background}
\label{sec:background}
In this section we introduce the relevant mathematical notation and provide background on Bayesian neural networks as well as the cold posterior effect.
\paragraph{Notation.} We study the standard supervised learning setting, where we have a dataset consisting of $n \in \mathbb{N}$ i.i.d. input-target pairs $\{(\bm{x}_i, {y}_i)\}_{i=1}^{n}$ distributed according to some (unknown) data distribution, $(\bm{x}_i, {y}_i) \sim \mathcal{D}$. We assume that the inputs come from some possibly high-dimensional space $\bm{x} \in \mathcal{X} \subset \mathbb{R}^{d}$ and that the targets are univariate, ${y} \in \mathbb{R}$. In a classification context, ${y}$ will for instance denote a binary encoding. All our results however extend to the multivariate target case. Occasionally we will find it useful to summarize inputs and targets into matrices, we will then denote $\bm{X} \in \mathbb{R}^{n \times d}$ and $\bm{y} \in \mathbb{R}^{n}$. We consider a family of functions $f_{\bm{\theta}}: \mathbb{R}^{d} \xrightarrow[]{} \mathbb{R}$ parameterized by $\bm{\theta} \in \mathbb{R}^{p}$, serving as models for the data distribution $\mathcal{D}$. 
\paragraph{BNNs.} Finding optimal values for $\bm{\theta}$ is a very challenging task due to the high-dimensional nature of the parameter space and its inherent degeneracy.  
A Bayesian approach to this problem specifies a \textit{prior} distribution $p(\bm{\theta})$ over the parameters (possibly incorporating domain knowledge) and leverages the data by means of Bayes' rule, resulting in the \textit{posterior} distribution
\begin{equation}
    \label{eq:posterior}
    p(\bm{\theta}|\bm{y}, \bm{X}) \propto p(\bm{y}|\bm{\theta},\bm{X})p(\bm{\theta})
\end{equation}

Prediction on a test sample $\bm{x}$ is then performed by marginalizing out the parameters $\bm{\theta}$ over the posterior distribution:
\begin{equation}
\label{eq:bma}
    p({y} | \bm{x}, \bm{X}, \bm{y}) = \int p({y} | \bm{x}, \bm{\theta}) p(\bm{\theta} | \bm{y}, \bm{X}) d\bm{\theta} .
\end{equation}
Eq.~\ref{eq:bma}, also called \emph{Bayesian model average} (BMA), naturally exploits the multi-modal landscape by averaging over all models that are compatible with the data, leading to richer explanations. Moreover, having access to the predictive distribution $p({y} | \bm{x}, \bm{X}, \bm{y})$ directly allows for uncertainty estimation. 
\paragraph{Inference.} In practice however, the integral in Eq. \ref{eq:bma} is intractable and as a consequence, approximate inference methods have been designed to estimate it. In particular, in this work we will make use of gradient-based Monte Carlo methods (MCMC), where a finite number of $K$ samples from the posterior are obtained and the BMA is approximated as $p({y} | \bm{x}, \bm{X}, \bm{y}) \approx \frac{1}{K}\sum_{k=1}^K p({y} | \bm{x}, \bm{\theta}_k)$, where $\bm{\theta}_k \sim p(\bm{\theta} | \bm{y}, \bm{X})$. To that end, we introduce the posterior energy function
$$U(\bm{\theta}) := -\log\left(p(\bm{\theta}|\bm{y}, \bm{X})\right),$$
along with the discretized Langevin dynamics which govern the parameters' evolution,

\begin{equation}
\label{eq:discr_langevin}
    \bm{\theta}_{t+1} \leftarrow \bm{\theta}_t - \frac{\alpha_t}{2}\nabla_{\bm{\theta}}U(\bm{\theta}) \\ 
     + \sqrt{\alpha_t}\mathcal{N}(0, \mathds{1}).
\end{equation}
In deep learning, we often deal with very big data corpora, rendering the gradient operation $\nabla_{\bm{\theta}}U(\bm{\theta})$ intractable. Inspired by the mini-batching operation in SGD, \citet{welling2011bayesian} introduced the same idea, crucially relying on the fact that for i.i.d. data $\{(\bm{x}_i, y_i)\}_{i=1}^{n}$, we can write
\begin{equation}
    \label{eq:posterior-energy}
    U(\bm{\theta}) \stackrel{i.i.d.}{=} -\sum_{i=1}^{n}\log\left(p({y}_i|\bm{\theta}, \bm{x}_i)\right) - \log(p(\bm{\theta})) .   
\end{equation}
A noisy estimate of the full gradient is then formed on a mini-batch $S_t \subset \{1, \dots, n\}$ of the data,
$$\nabla_{\bm{\theta}}U(\bm{\theta})\approx -\frac{n}{|S_t|} \nabla_{\bm{\theta}}\left(\sum_{i \in S_t} \log p(y_i |\bm{\theta}, \bm{x}_i) + \log p(\bm{\theta})\right).$$ 

These so-called SG-MCMC methods, in combination with pre-conditioners \citep{li2016preconditionedsgld} and cyclical learning rates \citep{zhang2019cyclicalsgmcmc}, offer a powerful and scalable approach to Bayesian inference, as shown in \citep{wenzel2020good}.
\paragraph{Cold Posteriors.} Despite their benefits, the adoption of BNNs is not widespread: inference procedures are usually slower and reported to even be outperformed by SGD in certain settings \citep{wenzel2020good}.
At the same time however, \citet{wenzel2020good} show that the performance issue can be resolved by re-scaling the posterior with a temperature parameter $T>0$:
\begin{equation}
    \label{eq:tempered_post}
    p(\bm{\theta} | \bm{y}, \bm{X})^{\frac{1}{T}} \propto p(\bm{y} | \bm{\theta}, \bm{X})^{\frac{1}{T}} p(\bm{\theta})^{\frac{1}{T}}.
\end{equation}
The optimal temperature is found to be consistently smaller than one, i.e. $T << 1$, across several models and datasets. Thus, the term "cold posterior effect" was coined. As discussed in Sec.~\ref{sec:intro}, exactly pin-pointing the origin of this effect is complicated and a number of hypotheses have been put forth in the literature. In this work, we focus on the role of data augmentation since empirically it is observed to be the main driver of the CPE \citep{noci2021disentangling, izmailov2021bayesian}.
Finally, we also consider the variant where only the likelihood is tempered, and not the prior:
\begin{equation}
    \label{eq:tempered_likelihood}
    p_T(\bm{\theta} | \bm{y}, \bm{X}) \propto p(\bm{y} | \bm{\theta}, \bm{X})^{\frac{1}{T}} p(\bm{\theta}) .
\end{equation}
We will show both theoretically and experimentally that tempering the likelihood is sufficient to account for the misspecification introduced by data augmentation. 

\paragraph{Data Augmentation.} A common technique in deep learning to foster invariance of a model is given by data augmentation. Given an example $(\bm{x}, {y}) \sim \mathcal{D}$, one produces several (random) augmentations of the input $\bm{x}$, that by design, should preserve the label information, i.e. ${{\tilde{y}}}={y}$. More formally, depending on the domain, the practitioner designs a (parametrized) augmentation function $R_{\bm{\eta}}:\mathbb{R}^{d} \xrightarrow[]{} \mathbb{R}^{d}$ that takes an input $\bm{x}$ and a parameter $\bm{\eta}\sim p({\bm{\eta}})$ and produces an augmented example ${\bm{\tilde{x}}} := R_{\bm{\eta}}(\bm{x})$. Usually, $R_{\bm{\eta}}$ is chosen in a way that the resulting augmentation ${\bm{\tilde{x}}}$ preserves the label, i.e. an annotator would assign the same label to ${\bm{\tilde{x}}}$ as $\bm{x}$. In a computer vision context, $R_{\bm{\eta}}$ would for instance correspond to the composition of randomly rotating, flipping and translating the input image, intuitively leaving the associated label invariant. The same input $\bm{x}$ is usually augmented several times, leading to a set of augmentations ${\bm{\tilde{x}}}_1, \dots, {\bm{\tilde{x}}}_B$. For instance, in stochastic gradient descent a fresh augmentation of $\bm{x}$ is produced at every epoch of the optimization. Data augmentation is a standard component of most deep learning pipelines and an almost necessary ingredient for state-of-the-art results. For instance, the top $5$ leaders in \textit{ImageNet} accuracy\footnote{\url{https://paperswithcode.com/sota/}}  \citep{dai2021coatnet, zhai2021scaling, 49959, liu2021swin, yuan2021florence}, all use some form of data augmentation.
\paragraph{Group Convolutions.}
In order to experimentally verify the predicted relationship between optimal temperature and model invariance, we make use of group-equivariant convolutions \cite{cohen2016group}, which extend the translation equivariance property of standard convolutions to richer classes of transformations ${G}$ that form finite symmetry groups. Given a feature map $\phi(\bm{x}) \in \mathbb{R}^{d}$ and a transformation $g:\mathbb{R}^{d} \xrightarrow[]{} \mathbb{R}^{d}$ for $g \in {G}$, we say that $\phi(\bm{x})$ is equivariant with respect to $G$ if $\exists\hspace{1mm} g' \in {G}$ s.t. $\phi(g(\bm{x})) = g'(\phi(\bm{x}))$. Note that invariance is the special case in which $g'$ is the identity function. 
The mathematical machinery of symmetry groups and equivariance can be combined to design group-equivariant convolutional layers. More concretely, given a feature map $\phi$ and $K$ filters $f_k$, where both the filters and feature map are functions on $G$ for all but the first layer, a  $G$-convolutional layer can be defined for $h \in G$, as: 
\begin{equation}
    [f * \phi](h) = \sum_{g \in {G}}\sum_{k=1}^{K}f_{k}(g)\phi(h^{-1}(g)) .
\end{equation}
For the first layer, $K$ is the number of channels of the input image, $f_k$ is the $k$-th channel of the input image, and the first sum is over its pixel values.

 Invariant architectures with respect to these groups can be built by using $G$-convolutional layers followed by a global average pooling layer (GAP) \citep{cohen2016group, veeling2018rotation}. However, one has to to be careful when designing such architectures, as some commonly used features break such invariances. In particular, it was shown that strided convolutions \citep{mouton2021stride}, pooling layers (subsampling) \citep{bulusu2021generalization} and padding \citep{kayhan2020translation} break translational equivariance in standard convolutions. 
 
\section{Data Augmentation in Bayesian Models}
\paragraph{Correlations.}
Data augmentation in the context of Bayesian inference is a delicate matter. 
\begin{figure}
    \centering
    \includegraphics[width=0.5\textwidth]{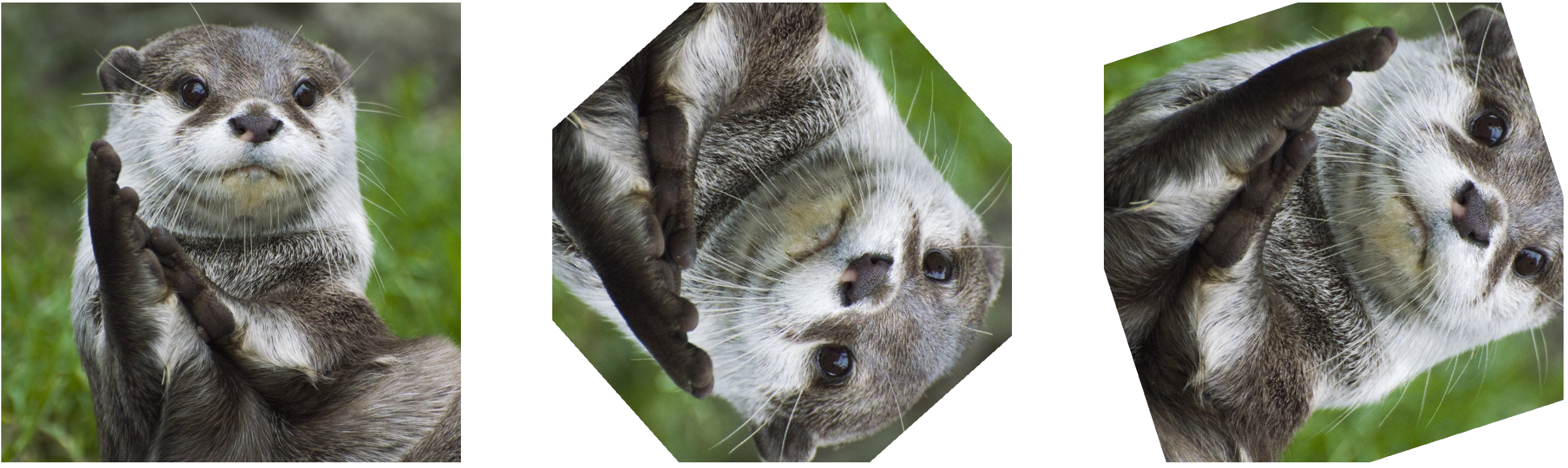}
    \caption{Illustration of augmentations, the original image $\bm{x}$ (left), and two random augmentation $R_{\bm{\eta}_1}(\bm{x})$ (middle) and  $R_{\bm{\eta}_2}(\bm{x})$ (right). We apply a composition of random rotations, crops and flips.}
    \label{fig:augmentations}
\end{figure}
Consider the three images in Fig.~\ref{fig:augmentations}, where the left image corresponds to the original sample $\bm{x}$ and the other two images $R_{\bm{\eta}_1}(\bm{x})$, $R_{\bm{\eta}_2}(\bm{x})$ are two random augmentations. The resulting label for the augmented samples is still ``otter", i.e. ${{\tilde{y}}}_1 = {{\tilde{y}}}_2 = {y}$ as the augmentations are chosen in a way to preserve that information. Very intuitively however, the two augmented examples share significant correlation. More concretely, considering $B \in \mathbb{N}$ augmentations of $\bm{x}$, forming the set $\tilde{\mathcal{S}}:=\{({\bm{\tilde{x}}}_1, {{{y}}}), \dots, ({\bm{\tilde{x}}}_B, {{{y}}})\}$, it is evident that the set exhibits significant correlation, leading to an effective sample size that lies between $1$ and $B$. However, the inference methods employed in the BNN literature implicitly assume i.i.d. data \citep{welling2011bayesian, NIPS2011_7eb3c8be, blundell2015weightuncertainty, wenzel2020good}, (essentially, boiling down to Eq.~\ref{eq:posterior-energy}), leading to a misspecified model when used in conjunction with data augmentation.  

\begin{figure*}
    \centering
    \begin{subfigure}[t]{0.47\textwidth}
        \includegraphics[width=\textwidth]{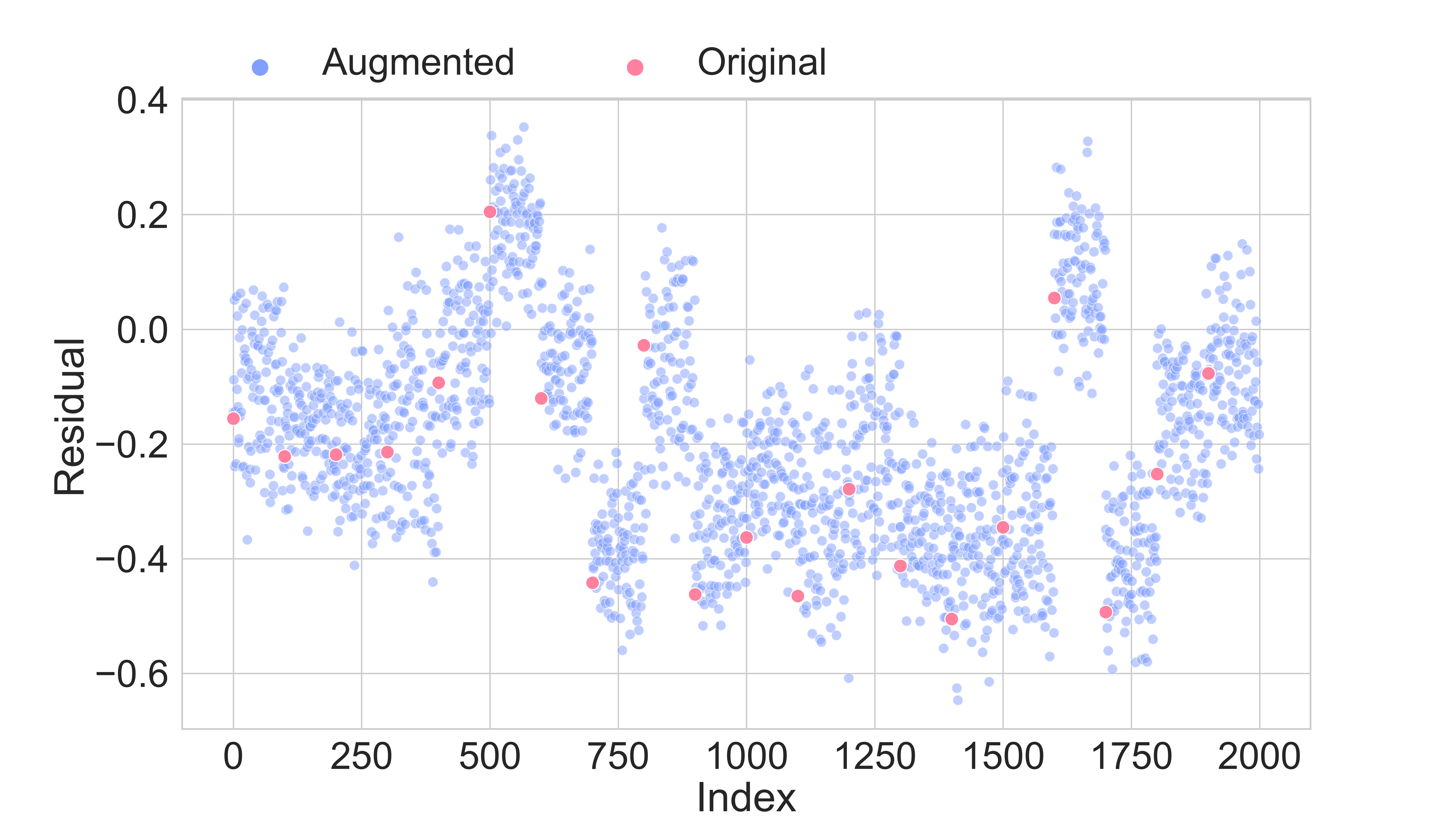}
        \caption{Augmented Residuals}
    \end{subfigure}
    \begin{subfigure}[t]{0.47\textwidth}
        \includegraphics[width=\textwidth]{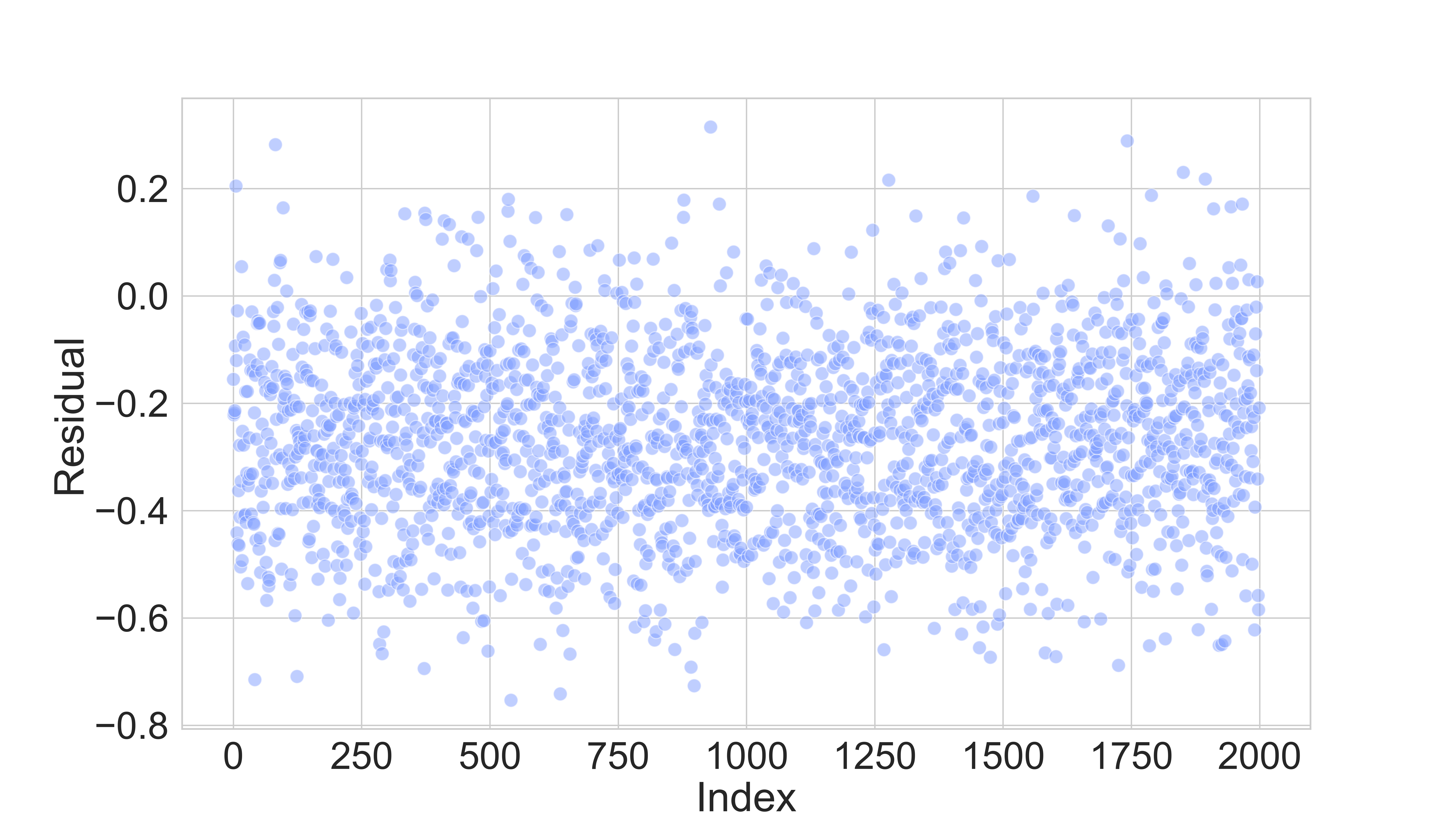}
        \caption{Unaugmented Residuals}
    \end{subfigure}
    \caption{\textbf{Residual vs. Order Plot:} Residuals of an untrained \textit{ResNet18} on \textit{Dogs vs Cats} dataset. (a) $20$ independent samples (red dots) of label ``cat" with $100$ augmentations each. (b) $2000$  independent samples of label ``cat". Augmented residuals display a strong dependence (forming clusters) while unaugmented residuals show no trend.}
    \label{fig:residuals}
\end{figure*}
\paragraph{A Simple Example.} We illustrate the arising correlations and the benefits of tempering through a very simple example. We study the classic textbook problem, where we aim to estimate the mean of a given set of samples $\{x_1, \dots, x_n\} \subset \mathbb{R}$. We will show in the next section how the intuition translates to a regression setting.  We make the modeling assumption
$$x|\mu \sim \mathcal{N}(\mu, \sigma^2),$$
where $\sigma>0$ is known but $\mu \in \mathbb{R}$ is unknown. Moreover, we set the prior $\mu \sim \mathcal{N}(\mu_0, \sigma^2_0)$ for $\sigma_0>0$ and $\mu_0 \in \mathbb{R}$. Given a sample $x$, we can augment it as $\tilde{x} = R_{\eta}(x)= x + \eta$ where $\eta \sim \mathcal{N}(0, \sigma_{\eta}^2)$ is independent from $x$, leaving the distribution invariant as evidently, $\tilde{x} \sim \mathcal{N}(\mu, \sigma^2+\sigma_{\eta}^2)$. On the other hand, a correlation structure among augmented samples emerges, as for two augmentations $\tilde{x}_i = x +\eta_i$ and $\tilde{x}_j =x+\eta_j$ with independent $\eta_i$, $\eta_j$, it holds
$$\operatorname{cov}(\tilde{x}_i, \tilde{x}_j) = \sigma^2.$$
If we consider $B \in \mathbb{N}$ augmentations per sample $x_i$ and collect them into a vector $\bm{\tilde{x}} \in \mathbb{R}^{Bn}$, it holds
$${\bm{\tilde{x}}} \sim \mathcal{N}(\mu \bm{1}_{Bn}, \bm{\tilde{\Sigma}}),$$
where $\bm{\tilde{\Sigma}} \in \mathbb{R}^{Bn \times Bn}$ is block-diagonal with blocks $\bm{\Sigma} = \sigma^2_{\eta} \mathds{1} + \sigma^2\bm{1}_B\bm{1}_B^{T} \in \mathbb{R}^{B \times B}$ and $\bm{1}_m \in \mathbb{R}^{m}$ denotes the all-ones vector. On the other hand, we can choose to ignore the correlation structure, treating all the samples as i.i.d. realizations, leading to the diagonal covariance $
(\sigma^2 + \sigma^2_{\eta})\mathds{1}$. Denote by $p(\mu|\bm{\tilde{x}};\bm{\tilde{\Sigma}})$ the posterior incorporating correlation and by $p_{T}(\mu|\bm{\tilde{x}};\mathds{1})$ the tempered posterior under the i.i.d. assumption. We can prove the following:
\begin{theorem}
\label{thm:correlated_means}
$p_{T}(\mu|\bm{\tilde{x}};\mathds{1})$ and $p(\mu|\bm{\tilde{x}};\bm{\tilde{\Sigma}})$  exactly match for the choice of temperature
$$T^{*}(\sigma_{\eta};B) = \frac{\sigma^2_{\eta}+ B\sigma^2}{\sigma^2_{\eta}+\sigma^2}.$$
\end{theorem}
We postpone the proof of Thm.~\ref{thm:correlated_means} to Appendix~\ref{proof:Thm}. Thm.~\ref{thm:correlated_means} shows that tempering completely fixes the misspecification implied by treating augmentations as i.i.d. samples! Let us comment on a few characteristics of the ideal temperature $T^{*}(\sigma_{\eta};B)$. First, it holds that $T^{*} \geq 1$, i.e. we always require a hot posterior. Moreover, $T^{*}$ is an increasing function in $B$, i.e. the more you augment, the hotter the posterior needs to become. This is intuitive, as we should rely less on the data since the sample size is artificially inflated. If a single augmentation is used, i.e. $B=1$ we recover $T^{*} = 1$ as expected, since data points indeed become independent. Finally $T^{*}$ is a decreasing function in $\sigma_{\eta}^2$, when augmentations become less diverse, i.e. $\sigma_{\eta} \xrightarrow[]{} 0$, we converge to $T^{*}=B$. Data augmentation is usually associated with cold posteriors in the literature, which is in stark contrast to our result. We explain this discrepancy in detail in Sec.~\ref{sec:exps}.

\begin{figure}
    \centering
    \includegraphics[width=0.33\textwidth]{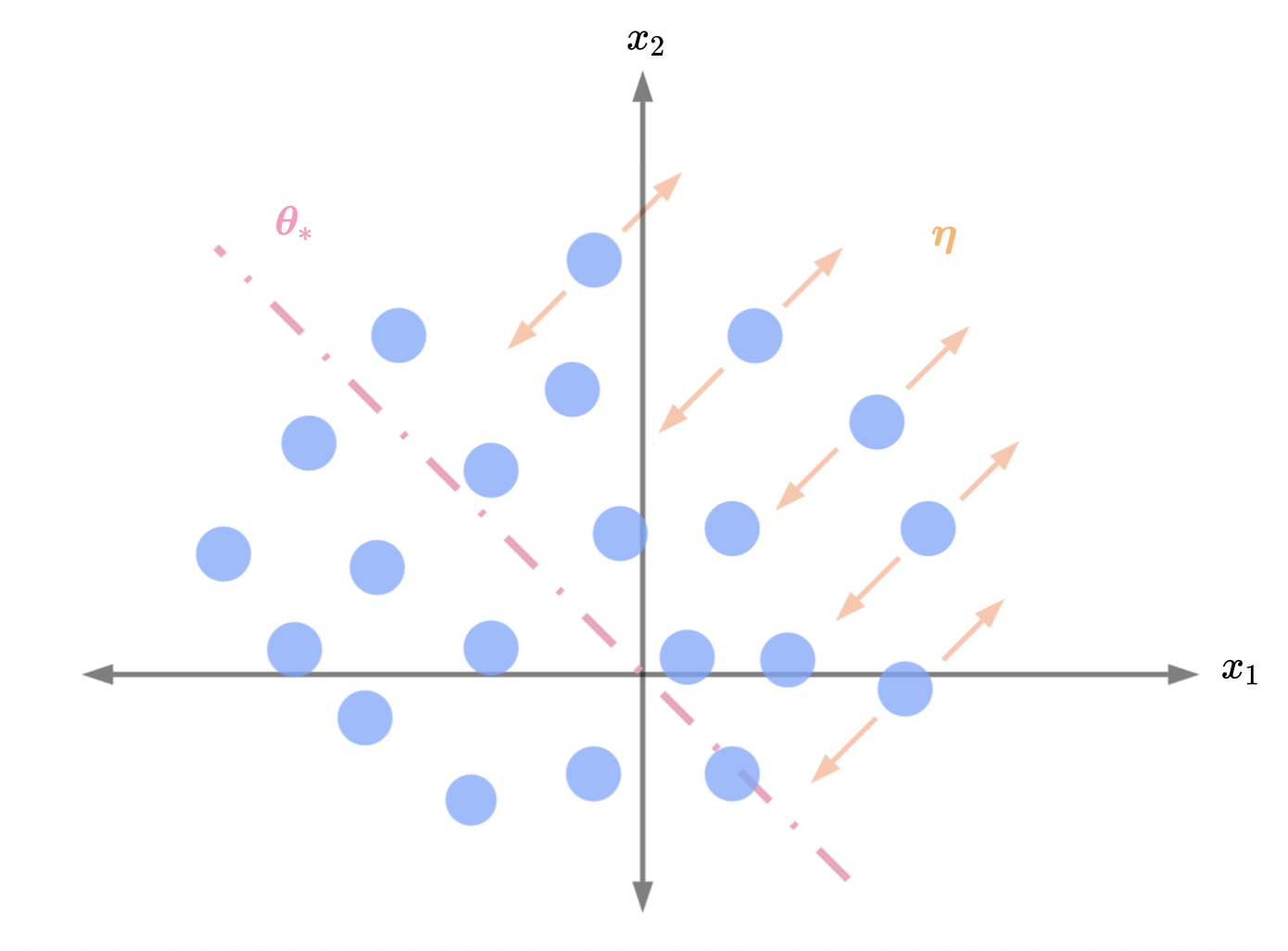}
    \caption{Illustration of additive augmentations $\bm{\eta}$ (orange) constrained to a line with an invariant model $\bm{\theta}_{*}$ (pink).}
    \label{fig:additive_augmentation}
\end{figure}

\paragraph{Regression Setting.}  Let us illustrate how a regression setting changes the implied model. Consider the data generating process 
$$y = f_{\bm{\theta}_{*}}(\bm{x}) + \epsilon ,$$
where $f_{\bm{\theta}}$ defines a family of functions, parametrized by $\bm{\theta}$ with true but unknown configuration $\bm{\theta}_{*} \in \mathbb{R}^{p}$. Moreover, $\bm{x} \in \mathbb{R}^{d}$ denotes the covariate and $\epsilon \sim \mathcal{N}(0, \sigma_{\epsilon}^2)$ is the error, inherent to the process. We place a prior $\bm{\theta} \sim p(\bm{\theta})$ on the parameters and assume that we have $n$ independent realizations of the process, $\{(\bm{x}_1, y_1), \dots, (\bm{x}_n, y_n)\}$. We consider general augmentations $R_{\bm{\eta}}(\bm{x})$ where $\bm{\eta} \sim p({\bm{\eta}})$ governs the randomness.
Similarly as in practice, since we cannot access $\bm{\theta}_{*}$, we postulate that the augmentation $\bm{\tilde{x}} := R_{\bm{\eta}}(\bm{x})$ shares the same response value, i.e. we form the sample $(\bm{\tilde{x}}, y)$. Imposing a response induces an error $\tilde{\epsilon}$ which we can calculate, due to the relation
$$y=f_{\bm{\theta}_{*}}(\bm{x}) + \epsilon \stackrel{!}{=} f_{\bm{\theta}_{*}}(R_{\bm{\eta}}(\bm{{x}})) + \tilde{\epsilon},$$
which after re-arranging, leads to the following:
$$\tilde{\epsilon} = \epsilon + \delta_{\bm{\eta}} ,$$
where we define $\delta_{\bm{\eta}} := f_{\bm{\theta}_*}(\bm{{x}}) - f_{\bm{\theta}_*}(R_{\bm{\eta}}(\bm{x}))$. Intuitively, $\delta_{\bm{\eta}}$ measures the degree of invariance of the true model $\bm{\theta}_{*}$. We make the natural assumption that $\mathbb{E}[\delta_{\bm{\eta}}] = 0$, implying that $\tilde{\epsilon}$ remains a centered random variable as the augmentation is not inducing any bias. In practice, the same data point $\bm{x}$ is augmented multiple times, i.e. we produce $R_{\bm{\eta}_i}(\bm{x})$ where $\bm{\eta}_i \stackrel{i.i.d.}{\sim} p(\bm{\eta})$ for $i=1, \dots, B$, which we associate with the same response $y$. This induces a series of errors $\tilde{\epsilon}_i$, which give rise to a correlation structure
\begin{equation}
    \label{eq:correlation}
    \begin{split}
        \operatorname{cor}(\tilde{\epsilon_i},\tilde{\epsilon_j}) = \frac{\sigma_{\epsilon}^2}{\sigma_{\epsilon}^2 + \operatorname{var}(\delta_{\bm{\eta}})}.
    \end{split}
\end{equation}
We detail the calculation in Appendix~\ref{app:correlations}.
This structure depends on the original error variance $\sigma_{\epsilon}^2$ and $\operatorname{var}(\delta_{\bm{\eta}})$, which in turn depends on the augmentation scheme $R$ and the true model $f_{\bm{\theta}_{*}}$. As a consequence, the likelihood \textbf{cannot} be factorized, i.e.
$$p(\tilde{y}_1, \dots, \tilde{y}_B|\bm{x}, \bm{\theta}) \not= \prod_{i=1}^{B}p(\tilde{y}_i|\bm{x}, \bm{\theta}).$$
We empirically demonstrate the correlated nature of the errors in Fig.~\ref{fig:residuals} by resorting to the classic tool of the \textit{Residual vs. Order} plot, often employed in least squares. We approximate the true (but unobservable) errors by the resulting residuals of a random \textit{ResNet18} for both augmented and non-augmented data. In the augmentation case, we observe a strong correlation when the residuals are plotted against the sample index, while without augmentations, they exhibit no structure. Notice that already a random (i.e. untrained) model displays this pattern, hinting at the fact that such invariances (and hence correlations) are built into the architecture. \\[2mm]
Finally our results also extend to the classification setting, we refer the reader to Appendix~\ref{app:logistic}.
\paragraph{Tempering as Effective Sample Size.} On the contrary, if errors become perfectly correlated, the likelihood even degenerates,
$$p(\tilde{y}_1, \dots, \tilde{y}_B|\bm{x}, \bm{\theta}) = p(\tilde{y}|\bm{x}, \bm{\theta})$$
We see that \textbf{wrongly} factorizing the likelihood in this case can be fixed through tempering with $T>0$, 
\begin{equation*}
    \begin{split}
        p(\tilde{y}_1, \dots, \tilde{y}_B|\bm{x}, \bm{\theta})^{\frac{1}{T}} &\stackrel{\text{wrong}}{=}  \prod_{i=1}^{B}p(\tilde{y}_i|\bm{x}, \bm{\theta})^{\frac{1}{T}}\\ &= p(\tilde{y}|\bm{x}, \bm{\theta})^{\frac{B}{T}}
    \end{split}
\end{equation*}
i.e. setting $T=B$ recovers the correct likelihood. Interpreting Eq.~ \ref{eq:correlation} as a measure of ``factorizability" of the likelihood, we conjecture that $T^{*} \in [1, B]$, depending on the degree of invariance of the model $\bm{\theta}_*$. In this sense, $\frac{B}{T}$ measures the ``effective sample size" of augmentations. As a consequence, there should ideally be a separate temperature $T_i$ for every datapoint $\bm{x}_i$. However, we conjecture that in realistic settings, augmentations of different datapoints should exhibit similar correlations, hence a single global temperature $T$ provides a good approximation. For intermediate values of correlations, tempering can reduce but not perfectly fix the misspecification. We refer to Appendix~\ref{app:inter-correlations} for details.
\paragraph{Linear Regression.} We can illustrate the general framework with the simpler case of linear regression, i.e. $f_{\bm{\theta}}(\bm{x}) = \bm{\theta}^{T}\bm{x}$ under additive augmentations $R_{\bm{\eta}}(\bm{x}) = \bm{x} + \bm{\eta}$ with $\bm{\eta} \sim \mathcal{N}(\bm{0}, \bm{\Sigma}_{\bm{\eta}})$. We can write 
\begin{equation*}
    \begin{split}
        \delta_{\bm{\eta}} = \bm{\theta}_{*}^{T}\bm{x} -\bm{\theta}_{*}^{T}R_{\bm{\eta}}(\bm{{x}}) = -\bm{\eta}^{T}\bm{\theta}_{*}
    \end{split}
\end{equation*}
Here, we can directly see that $\mathbb{E}[\delta_{\bm{\eta}}] = 0$. Moreover, the correlation of the errors can be computed in closed form as well (detailed in Appendix~\ref{app:linear_correlation}), 
\begin{equation*}
    \operatorname{cor}(\tilde{\epsilon_i},\tilde{\epsilon_j}) = \frac{\sigma^2_{\epsilon}}{\sigma^2_{\epsilon} + \bm{\theta}_{*}^{T}\bm{\Sigma}_{\bm{\eta}}\bm{\theta}_{*}}
\end{equation*}
We notice the following: If the true model $\bm{\theta}_{*}$ is contained in the null space of $\bm{\Sigma}_{\bm{\eta}}$, the errors exhibit perfect correlation, i.e. $\operatorname{cor}(\tilde{\epsilon_i},\tilde{\epsilon_j})=1$. This is very intuitive and illustrated in Fig.~\ref{fig:additive_augmentation} for $d=2$. If $\bm{\eta}$ is degenerate and only operates in a subspace, here along the $45 \degree$-axis e.g. $\bm{\Sigma}_{\bm{\eta}} = \begin{pmatrix} \sigma^2_{\eta} & \sigma^2_{\eta} \\
\sigma^2_{\eta} & \sigma^2_{\eta} \end{pmatrix}$, any orthogonal model $\bm{\theta}_{*}$ will be invariant to the augmentation, leading to perfectly correlated errors. On the other hand, once the augmentation noise $\sigma_{\eta}^2$ overwhelms the signal (i.e. $\sigma_{\eta}^2 \xrightarrow[]{} \infty$) we recover independent (but useless) samples. 

\paragraph{Conditioning or Not Conditioning.} In the previous section we have assumed, by conditioning on $\bm{x}$, that the model actually has access to $\bm{x}$. However, this is not what is done in practice, as the model makes inference relying only on the augmentations and not on the underlying datapoint. 
Now we derive a general generative model for data augmentation that does not rely on conditioning on the original datapoints $\bm{x}$, but only on the augmentations, and discuss its implications. 

Let $p(\tilde{\bm{x}}_i | \bm{x})$ be the likelihood of the augmentation $\tilde{\bm{x}}_i$ being generated from $\bm{x}$ through $R_{\bm{\eta}}(\bm{x})$. If we do not condition on $\bm{x}$, the likelihood for the augmentations $\tilde{\mathcal{S}}$ has the following form:
\begin{equation*}
\begin{split}
     p(\tilde{\mathcal{S}} | \bm{\theta}) &= \int p(\{({\bm{\tilde{x}}}_i, \tilde{{\bm{y}}}_i)\}_{i=1}^B, \bm{x}, | \bm{\theta}) d\bm{x}\\
    &= \int p(\{\tilde{\bm{y}}_i\}_{i=1}^B |\bm{x}, \{\bm{\tilde{x}}_i\}_{i=1}^B, \bm{\theta})\prod_{i=1}^B p(\tilde{\bm{x}}_i | \bm{x}) p(\bm{x}) d\bm{x} .
\end{split}
\end{equation*}
 Note that this intractable setting is what is commonly done in practice, as the model does not rely on $\bm{x}$, but only on the augmentations to estimate the posterior. This means that we have to marginalize $\bm{x}$ out in order to account for all possible original datapoints that might have generated $\tilde{\bm{x}}$. However, more realistically, it is safe to assume that an augmentation $\tilde{\bm{x}}$ is ``sufficiently close" to $\bm{x}$ that no other $\bm{x}'$ might have generated it. Therefore conditioning on $\bm{x}$ does not affect the parameter estimation. Clearly, this final reasoning assumes that the augmentations are not ``too wild". In particular, standard augmentations adopted in computer vision, such as rotations, horizontal/vertical flips and crops certainly satisfy this condition. 

\section{Experiments}
\label{sec:exps}
To perform approximate inference for BNNs, we use the SG-MCMC sampler described in \cite{wenzel2020good}, which includes cyclical step size \citep{zhang2019cyclicalsgmcmc} and layer-wise preconditioning (we refer the reader to Appendix~\ref{app:experiments} for the implementation details). 

\paragraph{Hot or Cold Posteriors?} To make the approximate inference method match the considered theoretical setting, we adjust the sample size from $n$ to $Bn$ where $B$ is the number of augmentations used for inference, which coincides with the number of training epochs. 
In order to only temper the likelihood, we apply a simple re-parameterization to the learning rate $\gamma_t = T\alpha_t$ and remove the temperature term from the prior, resulting in the overall update:
\begin{multline}
\label{eq:temp_likelihood_sgmcmc}
    \bm{\theta}_{t+1} \leftarrow \bm{\theta}_t - \frac{\gamma_t}{2}\left(\frac{Bn}{T |S_t|} \sum_{i \in S_t} \log p(y_i |  \bm{x}_i) + \log p(\bm{\theta}_t)\right) \\ 
     + \sqrt{\gamma_t}\mathcal{N}(0, \mathds{1}) ,
\end{multline}
 From Eq. \ref{eq:temp_likelihood_sgmcmc}, the role of $T$ in data augmentation is manifest: it rescales the augmented dataset size $Bn$ to adjust for the non i.i.d data. Inspired by our theoretical results, we conjecture that $T^* \in [1, B]$, resulting thus in a \emph{hot} posterior. On the contrary, if one does not explicitly account for the augmentations and considers only $n$ datapoints, then $T' = \frac{T^*}{B}$ is optimal, and \emph{cold} posteriors are obtained instead. In other words, hot and cold posteriors are two sides of the same coin in our setting, they are simply a consequence of the normalization used in SG-MCMC. In our experiment, we will adopt the parametrization of Eq.~\ref{eq:temp_likelihood_sgmcmc} to remain consistent with our theoretical results. Therefore hot posteriors are to be expected, and $T=B$ is optimal when either the model is invariant with respect to the augmentations, or data augmentation is switched off. In all the following plots, $T=B$ will be highlighted by a vertical dashed line.

\begin{figure}[h!]
\centering
\includegraphics[width=0.45\textwidth]{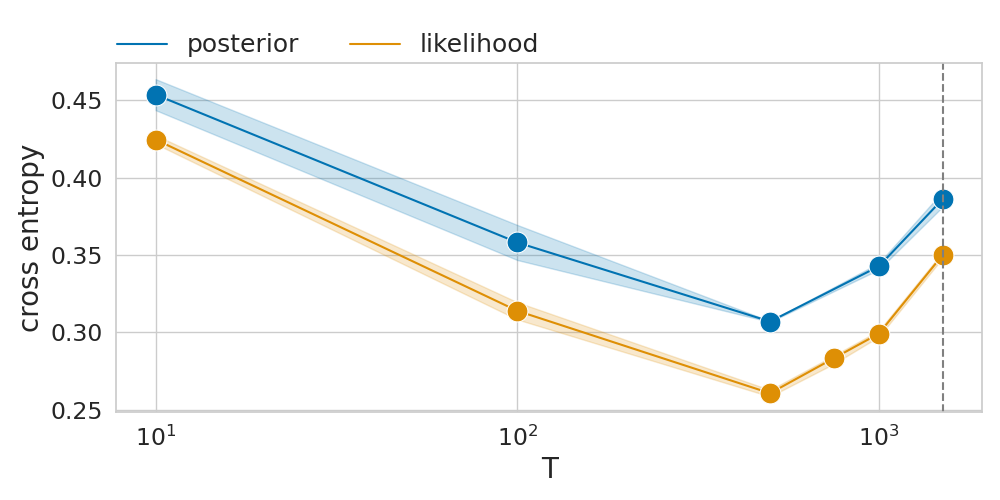}
\caption{Test cross-entropy as a function of the temperature $T$ for posterior tempering (blue) and likelihood tempering (orange).} 
\label{fig:lik_vs_posterior_temp}
\end{figure}

As a first experiment, we test whether tempering only the likelihood qualitatively changes the temperature landscape, compared to tempering the posterior. We display the results 
in Fig.~\ref{fig:lik_vs_posterior_temp} for a \textit{ResNet20} when data augmentation is used (random flips and crops). We observe that the same optimal temperature is achieved by both approaches, strongly suggesting that the likelihood is at the core of the CPE for data augmentation, and not the prior over the parameters, which was shown to play a significant role for small sample sizes \citep{noci2021disentangling}. Moreover, we indeed observe hot posteriors (i.e. $T>>1$) when employing the update in Eq.~\ref{eq:temp_likelihood_sgmcmc}, as predicted by our results.

\paragraph{G-Convolutions Alleviate the CPE.} As detailed in Section \ref{sec:background}, the adoption of $G$-convolutions is not sufficient to obtain an invariant architecture, as standard operations like striding or padding break the invariance. This is concerning, as strided convolutions are heavily used in state-of-the arts architectures like ResNets to enhance performance and reduce the parameter count. In our experiments, we use a \textit{ResNet20} with $2$-strided $G$-convolutions, thus losing its exact invariance but remaining largely insensitive to the group transformations. We will refer to it in the following as $G$-\textit{ResNet}. In particular, we will use \emph{p4m} equivariant convolutions, enforcing equivariance with respect to compositions of translations,
horizontal and vertical flip as well as rotations by multiples of 90$\degree$. Finally, note that \emph{p4m} convolutions effectively make the size of the feature maps 8 times larger. Therefore, to have a fair comparison between standard and group convolutions, we follow \citet{cohen2016group} and reduce the number of filters of $G$-convolutional layers by $\sqrt{8}$ to roughly have the same number of training parameters.

\begin{figure}[h!]
\centering
\includegraphics[width=0.45\textwidth]{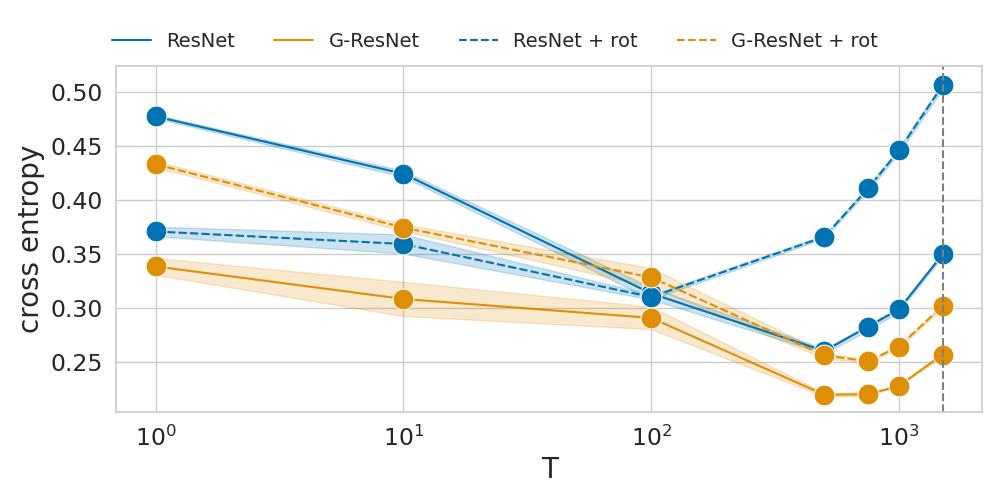}
\caption{Test cross-entropy as a function of the temperature $T$ for a ResNet with standard convolutions (blue) and G-convolutions (orange), with or without extra 90\degree  rotations (dashed lines).}
\label{fig:gconv_vs_conv}
\end{figure}

The results of this comparison are shown 
in Fig. \ref{fig:gconv_vs_conv}. First, we use only random crops and horizontal flips to augment the dataset (solid lines). Then we repeat the experiment with the additions of multiples of 90$\degree$ rotations (dashed lines). 
$G$-\textit{ResNet} outperforms \textit{ResNet20} at the optimal temperature in all cases, while 90$\degree$ rotations seems to degrade performance across both architectures. Note when using only flips and crops (solid lines), the optimal temperature is similar. 
However, when rotations are added, the optimal temperature of \textit{ResNet20} decreases significantly, while for $G$-\textit{ResNet} - which is more insensitive to such rotations - there is almost no shift in the optimal temperature.  In particular, we report a significant temperature shift in \textit{ResNet20} while almost no shift for $G$-\textit{ResNet}, confirming our hypothesis on the role of invariance. We stress
that the $G$-\textit{ResNet} is not invariant, but only \emph{more} invariant (insensitive), due to the usage of strides and random crops. Therefore we should not expect $T=B$ to be optimal in this case. 

\begin{figure}[h!]
\centering
\includegraphics[width=0.45\textwidth]{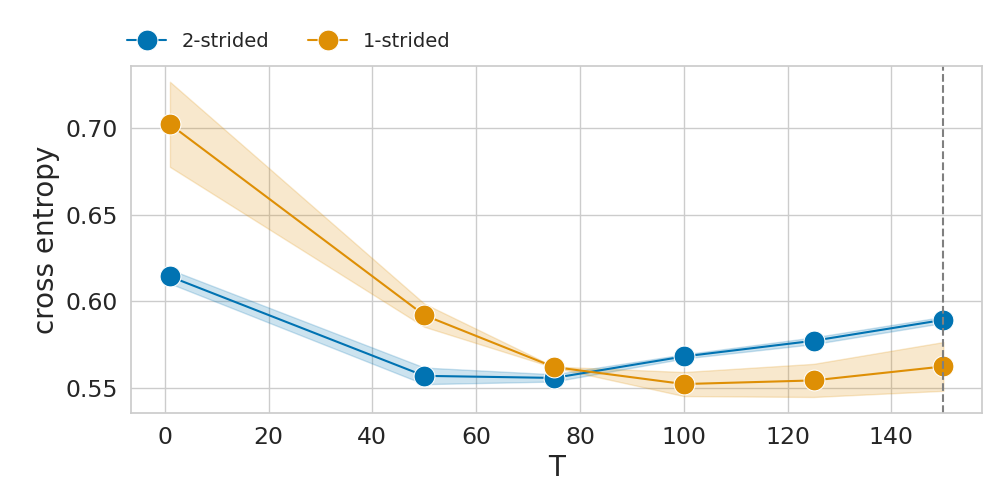}
\caption{Test cross-entropy as a function of the temperature for 2-strided (blue) and 1-strided (orange) G-convolutional network. Note how the 1-strided model presents a very flat curve toward $T=150$, indicating its insensitivity with respect to the augmentations. }
\label{fig:gconv_inv}
\end{figure}

\begin{figure}[h!]
\centering
\includegraphics[width=0.45\textwidth]{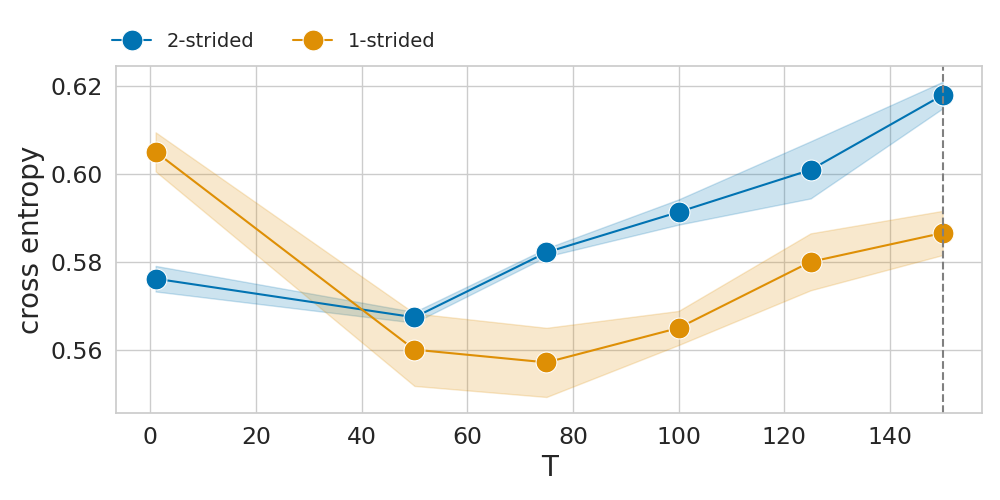}
\caption{Same setting as in Fig. \ref{fig:gconv_inv}, but this time we perform an extra random rotation of 10\degree. Note how the loss of invariance shifts the optimal temperature significantly, in both the 2-strided and 1-strided case.}
\label{fig:gconv_small_rot}
\end{figure}

\paragraph{An Almost Invariant Model.}
Next, we test whether an almost invariant model (\emph{insensitive}) can achieve an optimal temperature of $T=B$ while employing data augmentation. In our case, such a model can be built by \textbf{not} using strides (i.e. stride=1), and use only flips and multiples of 90$\degree$ rotations as augmentations. We fix the maximum number of augmentations to $B=150$, and use a burn-in period for SG-MCMC of 150 epochs. In this way, we do not start sampling before the model has visited all the datapoints. We display the results in Fig.~\ref{fig:gconv_inv}. As predicted, difference in performance at the optimal temperature and at $T=B=150$ is minimal. Further evidence for our argument is given by the fact that $2$-strided convolutions (i.e. a less invariant model) produce a weaker hot posterior effect. The invariance can be easily destroyed by adding an extra random 10$\degree$ rotation to the set of transformations, as shown in Fig. \ref{fig:gconv_small_rot}. In that case we move away from $T=150$, regardless of the stride.

\begin{figure}[h!]
\centering
\includegraphics[width=0.45\textwidth]{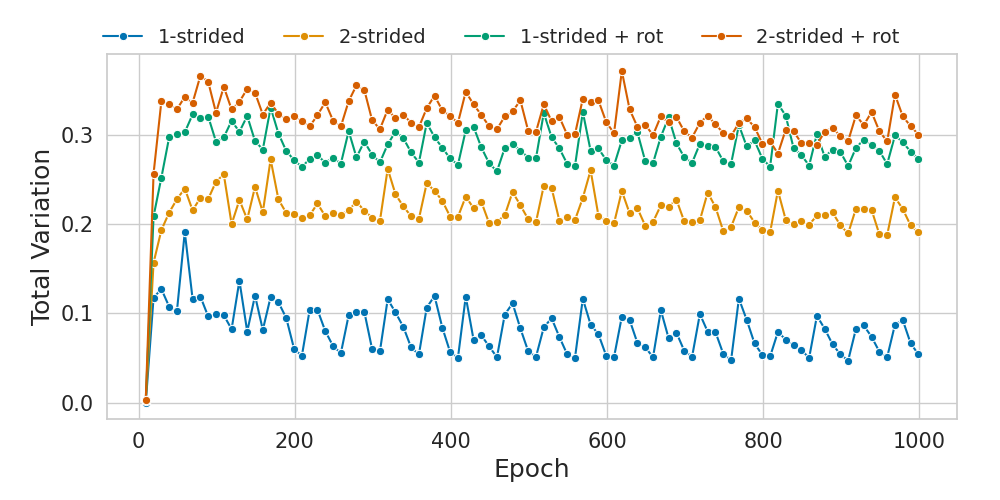}
\caption{Total variation during learning for the experiments with G-convolutional networks. A lower value of the total variation indicates a greater degree of invariance.}
\label{fig:inv_plot}
\end{figure}

Finally, to have a quantitative measure of the degree of invariance, we plot the absolute value of the difference of the output probabilities of the model (i.e the total variation) among the augmentations used for training. We use the models with optimal temperatures determined in the previous experiments (Fig. \ref{fig:gconv_inv}-\ref{fig:gconv_small_rot}). Results are shown in Fig.~\ref{fig:inv_plot}. As expected, we find that the total variation monotonically increases with the degree of invariance of model, i.e. the most invariant model ($1$-strided, no $10\degree$ rotations) displays the smallest variation. 


\section{Related Work}
\paragraph{Correlations.} Incorporating dependence between samples has a long tradition in machine learning. Correlations emerge naturally in a range of statistical applications, including longitudinal data \citep{longitud_cor}, time series \citep{Hamilton2020} and clustering \citep{bayesian_book}, to name but a few. The most relevant setting to our work is longitudinal modeling, where multiple measurements are produced from a single source point, a very common scenario in the field of biostatistics. Quantifying the predictive relationship between blood pressure and Diabetes (binary response) is a standard example. A typical dataset consists of $n$ patients, where crucially, for each patient $i$, repeated blood pressure measurements are performed due to the inherently noisy measurement process. Naturally, these repeated measurements are far from independent. Neglecting these correlations can be detrimental and as a consequence, various approaches have been developed, dating back to f\citet{longitud_cor, longitud_cor_2, Laird1982RandomeffectsMF}. We refer to \citet{book} for an overview. Many of those approaches build upon the well-known generalized least squares method or the linear mixed model \citep{linear_mixed}. 

\paragraph{Data Augmentation and CPE.}
While some works have explored data augmentation from a theoretical angle for standard neural networks \citep{data_augmentation, pmlr-v97-dao19b, Wu2020OnTG}, to the best of our knowledge only \citet{nabarro2021data} have explored it in the context of the cold posterior effect. They construct an invariant model by marginalizing over the augmented data distribution through averaging the predicted logits/probabilities of the model. They are however not able to explain the cold posterior effect. We argue that this is due to the fact that the augmentations are considered as latent variables that generated the observed (un-augmented) datapoints. Orthogonally to their approach, in this work we argue for the more realistic case in which the augmentations are generated from the original datapoints, and cold/hot posteriors arise from not taking into account the correlations between the errors that this process inevitably generates. This way, infinitely many augmentations can be considered without overwhelming the prior, in contrast to the argument in \citet{nabarro2021data}.
\paragraph{Model Misspecification and Tempering.}
Our work is inspired by a remarkable series of works on the role of tempering under model misspecification \cite{grunwald2011safe, grunwald2012safe, grunwald2017inconsistency}. In particular, in \citet{grunwald2017inconsistency} it is shown that wrongly modeling heteroscedastic noise errors as homoscedastic, makes Bayesian inference fail, while tempering fixes this model misspecification. In our case, the model misspecification is caused by not modeling the strong correlations arising from the data augmentation process. Finally, although relatively new, the CPE has been analyzed under other perspectives, and its flourishing literature includes other works such as \cite{zeno2020cold, adlam2020cold, laves2021posterior}.
\section{Conclusion}
In this work, we showed how data augmentation, in conjunction with invariance, introduces correlations between errors, leading to misspecified models. We demonstrated how tempering can reduce this misspecification by approximating the correct posterior, offering a possible explanation for the CPE. Tempering is thus more principled from a Bayesian perspective than previously assumed. We also identified $G$-convolutions to be a viable tool for the design of BNNs, enhancing their invariance and leading to better priors. Our theoretical and empirical results suggest several avenues towards combatting the CPE and improving BNNs in general.
From Figure \ref{fig:inv_plot}, it is manifest that the average invariance, as measured by the total variation, is determined during the burn-in epochs. Therefore, we foresee as a promising future direction to form an estimator of the ideal temperature that could be computed from the data during the burn-in period. Crucially, such an estimator could remove the need for expensive grid searches. Finally, the superior performances achieved by the models employing $G$-convolutions further motivates the developments of more informed priors over functions that have the desired invariances incorporated into the model. Finally, we want to highlight that the CPE has been shown to arise from other causes as well, which do not involve data augmentation \cite{fortuin2021bayesian, noci2021disentangling}. It remains exciting future work to complete the understanding of the CPE in these settings.

\bibliography{example_paper}

\begin{thebibliography}{50}
\providecommand{\natexlab}[1]{#1}
\providecommand{\url}[1]{\texttt{#1}}
\expandafter\ifx\csname urlstyle\endcsname\relax
  \providecommand{\doi}[1]{doi: #1}\else
  \providecommand{\doi}{doi: \begingroup \urlstyle{rm}\Url}\fi

\bibitem[Adlam et~al.(2020)Adlam, Snoek, and Smith]{adlam2020cold}
Adlam, B., Snoek, J., and Smith, S.~L.
\newblock Cold posteriors and aleatoric uncertainty.
\newblock \emph{arXiv preprint arXiv:2008.00029}, 2020.

\bibitem[Aitchison(2021)]{aitchison2020statistical}
Aitchison, L.
\newblock A statistical theory of cold posteriors in deep neural networks.
\newblock In \emph{International Conference on Learning Representations}, 2021.

\bibitem[Blundell et~al.(2015)Blundell, Cornebise, Kavukcuoglu, and
  Wierstra]{blundell2015weightuncertainty}
Blundell, C., Cornebise, J., Kavukcuoglu, K., and Wierstra, D.
\newblock In \emph{Proceedings of the 32nd International Conference on Machine
  Learning (ICML)}, volume~37, pp.\  1613--1622, 2015.

\bibitem[Bulusu et~al.(2021)Bulusu, Favoni, Ipp, M{\"u}ller, and
  Schuh]{bulusu2021generalization}
Bulusu, S., Favoni, M., Ipp, A., M{\"u}ller, D.~I., and Schuh, D.
\newblock Generalization capabilities of translationally equivariant neural
  networks.
\newblock \emph{arXiv preprint arXiv:2103.14686}, 2021.

\bibitem[Chen et~al.(2020)Chen, Dobriban, and Lee]{data_augmentation}
Chen, S., Dobriban, E., and Lee, J.
\newblock A group-theoretic framework for data augmentation.
\newblock In Larochelle, H., Ranzato, M., Hadsell, R., Balcan, M.~F., and Lin,
  H. (eds.), \emph{Advances in Neural Information Processing Systems},
  volume~33, pp.\  21321--21333. Curran Associates, Inc., 2020.

\bibitem[Cohen \& Welling(2016)Cohen and Welling]{cohen2016group}
Cohen, T. and Welling, M.
\newblock Group equivariant convolutional networks.
\newblock In \emph{International Conference on Machine Learning}, pp.\
  2990--2999, 2016.

\bibitem[Dai et~al.(2021)Dai, Liu, Le, and Tan]{dai2021coatnet}
Dai, Z., Liu, H., Le, Q.~V., and Tan, M.
\newblock Coatnet: Marrying convolution and attention for all data sizes.
\newblock In Beygelzimer, A., Dauphin, Y., Liang, P., and Vaughan, J.~W.
  (eds.), \emph{Advances in Neural Information Processing Systems}, 2021.

\bibitem[Dao et~al.(2019)Dao, Gu, Ratner, Smith, De~Sa, and
  Re]{pmlr-v97-dao19b}
Dao, T., Gu, A., Ratner, A., Smith, V., De~Sa, C., and Re, C.
\newblock A kernel theory of modern data augmentation.
\newblock In Chaudhuri, K. and Salakhutdinov, R. (eds.), \emph{Proceedings of
  the 36th International Conference on Machine Learning}, volume~97 of
  \emph{Proceedings of Machine Learning Research}, pp.\  1528--1537. PMLR,
  09--15 Jun 2019.

\bibitem[Devlin et~al.(2019)Devlin, Chang, Lee, and Toutanova]{devlin2019bert}
Devlin, J., Chang, M.-W., Lee, K., and Toutanova, K.
\newblock Bert: Pre-training of deep bidirectional transformers for language
  understanding.
\newblock \emph{Proceedings of NAACL-HLT}, 2019.

\bibitem[Draxler et~al.(2018)Draxler, Veschgini, Salmhofer, and
  Hamprecht]{landscape}
Draxler, F., Veschgini, K., Salmhofer, M., and Hamprecht, F.~A.
\newblock Essentially no barriers in neural network energy landscape.
\newblock In Dy, J.~G. and Krause, A. (eds.), \emph{Proceedings of the 35th
  International Conference on Machine Learning, {ICML} 2018,
  Stockholmsm{\"{a}}ssan, Stockholm, Sweden, July 10-15, 2018}, volume~80 of
  \emph{Proceedings of Machine Learning Research}, pp.\  1308--1317. {PMLR},
  2018.

\bibitem[Fortuin et~al.(2021)Fortuin, Garriga-Alonso, Wenzel, R{\"a}tsch,
  Turner, van~der Wilk, and Aitchison]{fortuin2021bayesian}
Fortuin, V., Garriga-Alonso, A., Wenzel, F., R{\"a}tsch, G., Turner, R.,
  van~der Wilk, M., and Aitchison, L.
\newblock Bayesian neural network priors revisited.
\newblock \emph{arXiv preprint arXiv:2102.06571}, 2021.

\bibitem[Garipov et~al.(2018)Garipov, Izmailov, Podoprikhin, Vetrov, and
  Wilson]{NEURIPS2018_be3087e7}
Garipov, T., Izmailov, P., Podoprikhin, D., Vetrov, D.~P., and Wilson, A.~G.
\newblock Loss surfaces, mode connectivity, and fast ensembling of dnns.
\newblock In Bengio, S., Wallach, H., Larochelle, H., Grauman, K.,
  Cesa-Bianchi, N., and Garnett, R. (eds.), \emph{Advances in Neural
  Information Processing Systems}, volume~31. Curran Associates, Inc., 2018.

\bibitem[Graves(2011{\natexlab{a}})]{NIPS2011_7eb3c8be}
Graves, A.
\newblock Practical variational inference for neural networks.
\newblock In Shawe-Taylor, J., Zemel, R., Bartlett, P., Pereira, F., and
  Weinberger, K. (eds.), \emph{Advances in Neural Information Processing
  Systems}, volume~24. Curran Associates, Inc., 2011{\natexlab{a}}.
\newblock URL
  \url{https://proceedings.neurips.cc/paper/2011/file/7eb3c8be3d411e8ebfab08eba5f49632-Paper.pdf}.

\bibitem[Graves(2011{\natexlab{b}})]{graves2011practicalvb}
Graves, A.
\newblock Practical variational inference for neural networks.
\newblock In \emph{Advances in neural information processing systems}, pp.\
  2348--2356, 2011{\natexlab{b}}.

\bibitem[Gr{\"u}nwald(2011)]{grunwald2011safe}
Gr{\"u}nwald, P.
\newblock Safe learning: bridging the gap between bayes, mdl and statistical
  learning theory via empirical convexity.
\newblock In \emph{Proceedings of the 24th Annual Conference on Learning
  Theory}, pp.\  397--420. JMLR Workshop and Conference Proceedings, 2011.

\bibitem[Gr{\"u}nwald(2012)]{grunwald2012safe}
Gr{\"u}nwald, P.
\newblock The safe bayesian.
\newblock In \emph{International Conference on Algorithmic Learning Theory},
  pp.\  169--183. Springer, 2012.

\bibitem[Gr{\"u}nwald et~al.(2017)Gr{\"u}nwald, Van~Ommen,
  et~al.]{grunwald2017inconsistency}
Gr{\"u}nwald, P., Van~Ommen, T., et~al.
\newblock Inconsistency of bayesian inference for misspecified linear models,
  and a proposal for repairing it.
\newblock \emph{Bayesian Analysis}, 12\penalty0 (4):\penalty0 1069--1103, 2017.

\bibitem[Hamilton(2020)]{Hamilton2020}
Hamilton, J.~D.
\newblock \emph{Time Series Analysis}.
\newblock Princeton University Press, 2020.
\newblock ISBN 9780691218632.
\newblock \doi{10.1515/9780691218632}.

\bibitem[He et~al.(2016)He, Zhang, Ren, and Sun]{resnet2016}
He, K., Zhang, X., Ren, S., and Sun, J.
\newblock Deep residual learning for image recognition.
\newblock In \emph{In IEEE Conference on Computer Vision and Pattern
  Recognition (CVPR)}, pp.\  770--778, 2016.

\bibitem[Hern{\'a}ndez-Lobato \& Adams(2015)Hern{\'a}ndez-Lobato and
  Adams]{hernandez2015probabilisticbp}
Hern{\'a}ndez-Lobato, J.~M. and Adams, R.
\newblock Probabilistic backpropagation for scalable learning of {Bayesian}
  neural networks.
\newblock In \emph{International Conference on Machine Learning}, pp.\
  1861--1869, 2015.

\bibitem[Izmailov et~al.(2021)Izmailov, Vikram, Hoffman, and
  Wilson]{izmailov2021bayesian}
Izmailov, P., Vikram, S., Hoffman, M.~D., and Wilson, A.~G.
\newblock What are bayesian neural network posteriors really like?
\newblock \emph{Proceedings of the 38th international conference on machine
  learning (ICML)}, 2021.

\bibitem[Jaynes(2003)]{jaynes_2003}
Jaynes, E.~T.
\newblock \emph{Probability Theory: The Logic of Science}.
\newblock Cambridge University Press, 2003.
\newblock \doi{10.1017/CBO9780511790423}.

\bibitem[Kayhan \& Gemert(2020)Kayhan and Gemert]{kayhan2020translation}
Kayhan, O.~S. and Gemert, J. C.~v.
\newblock On translation invariance in cnns: Convolutional layers can exploit
  absolute spatial location.
\newblock In \emph{Proceedings of the IEEE/CVF Conference on Computer Vision
  and Pattern Recognition}, pp.\  14274--14285, 2020.

\bibitem[Kolmogorov(1960)]{kolmogorov1960foundations}
Kolmogorov, A.~N.
\newblock \emph{Foundations of the Theory of Probability}.
\newblock Chelsea Pub Co, June 1960.

\bibitem[Laird \& Ware(1982)Laird and Ware]{Laird1982RandomeffectsMF}
Laird, N.~M. and Ware, J.~H.
\newblock Random-effects models for longitudinal data.
\newblock \emph{Biometrics}, 38 4:\penalty0 963--74, 1982.

\bibitem[Laves et~al.(2021)Laves, T{\"o}lle, Schlaefer, and
  Engelhardt]{laves2021posterior}
Laves, M.-H., T{\"o}lle, M., Schlaefer, A., and Engelhardt, S.
\newblock Posterior temperature optimization in variational inference for
  inverse problems.
\newblock \emph{arXiv preprint arXiv:2106.07533}, 2021.

\bibitem[Li et~al.(2016)Li, Chen, Carlson, and Carin]{li2016preconditionedsgld}
Li, C., Chen, C., Carlson, D., and Carin, L.
\newblock Preconditioned stochastic gradient {Langevin} dynamics for deep
  neural networks.
\newblock In \emph{Thirtieth AAAI Conference on Artificial Intelligence}, 2016.

\bibitem[Liang \& Zeger(1986)Liang and Zeger]{longitud_cor}
Liang, K.-Y. and Zeger, S.~L.
\newblock Longitudinal data analysis using generalized linear models.
\newblock \emph{Biometrika}, 73\penalty0 (1):\penalty0 13--22, 04 1986.
\newblock ISSN 0006-3444.

\bibitem[Liu et~al.(2021)Liu, Hu, Lin, Yao, Xie, Wei, Ning, Cao, Zhang, Dong,
  Wei, and Guo]{liu2021swin}
Liu, Z., Hu, H., Lin, Y., Yao, Z., Xie, Z., Wei, Y., Ning, J., Cao, Y., Zhang,
  Z., Dong, L., Wei, F., and Guo, B.
\newblock Swin transformer v2: Scaling up capacity and resolution, 2021.

\bibitem[Mouton et~al.(2021)Mouton, Myburgh, and Davel]{mouton2021stride}
Mouton, C., Myburgh, J.~C., and Davel, M.~H.
\newblock Stride and translation invariance in cnns.
\newblock In \emph{Southern African Conference for Artificial Intelligence
  Research}, pp.\  267--281. Springer, 2021.

\bibitem[Nabarro et~al.(2021)Nabarro, Ganev, Garriga-Alonso, Fortuin, van~der
  Wilk, and Aitchison]{nabarro2021data}
Nabarro, S., Ganev, S., Garriga-Alonso, A., Fortuin, V., van~der Wilk, M., and
  Aitchison, L.
\newblock Data augmentation in bayesian neural networks and the cold posterior
  effect, 2021.

\bibitem[Nguyen et~al.(2015)Nguyen, Yosinski, and Clune]{Nguyen2015DeepNN}
Nguyen, A.~M., Yosinski, J., and Clune, J.
\newblock Deep neural networks are easily fooled: High confidence predictions
  for unrecognizable images.
\newblock \emph{2015 IEEE Conference on Computer Vision and Pattern Recognition
  (CVPR)}, pp.\  427--436, 2015.

\bibitem[Noci et~al.(2021)Noci, Roth, Bachmann, Nowozin, and
  Hofmann]{noci2021disentangling}
Noci, L., Roth, K., Bachmann, G., Nowozin, S., and Hofmann, T.
\newblock Disentangling the roles of curation, data-augmentation and the prior
  in the cold posterior effect.
\newblock In \emph{Advances in Neural Information Processing Systems}, 2021.

\bibitem[Pham et~al.(2021)Pham, Dai, Xie, Luong, and Le]{49959}
Pham, H., Dai, Z., Xie, Q., Luong, M.-T., and Le, Q.~V.
\newblock Meta pseudo labels.
\newblock In \emph{IEEE Conference on Computer Vision and Pattern Recognition},
  2021.

\bibitem[Robinson(1991)]{linear_mixed}
Robinson, G.
\newblock That blup is a good thing: The estimation of random effects.
\newblock \emph{Statistical Science}, 6, 02 1991.
\newblock \doi{10.1214/ss/1177011926}.

\bibitem[Savage(1954)]{Savage1954}
Savage, L.~J.
\newblock \emph{The Foundations of Statistics}.
\newblock Wiley Publications in Statistics, 1954.

\bibitem[Silver et~al.(2016)Silver, Huang, Maddison, Guez, Sifre, Driessche,
  Schrittwieser, Antonoglou, Panneershelvam, Lanctot, Dieleman, Grewe, Nham,
  Kalchbrenner, Sutskever, Lillicrap, Leach, Kavukcuoglu, Graepel, and
  Hassabis]{alphago}
Silver, D., Huang, A., Maddison, C., Guez, A., Sifre, L., Driessche, G.,
  Schrittwieser, J., Antonoglou, I., Panneershelvam, V., Lanctot, M., Dieleman,
  S., Grewe, D., Nham, J., Kalchbrenner, N., Sutskever, I., Lillicrap, T.,
  Leach, M., Kavukcuoglu, K., Graepel, T., and Hassabis, D.
\newblock Mastering the game of go with deep neural networks and tree search.
\newblock \emph{Nature}, 529:\penalty0 484--489, 01 2016.
\newblock \doi{10.1038/nature16961}.

\bibitem[Szegedy et~al.(2013)Szegedy, Zaremba, Sutskever, Bruna, Erhan,
  Goodfellow, and Fergus]{szegedy2013intriguing}
Szegedy, C., Zaremba, W., Sutskever, I., Bruna, J., Erhan, D., Goodfellow, I.,
  and Fergus, R.
\newblock Intriguing properties of neural networks.
\newblock \emph{arXiv preprint arXiv:1312.6199}, 2013.

\bibitem[Veeling et~al.(2018)Veeling, Linmans, Winkens, Cohen, and
  Welling]{veeling2018rotation}
Veeling, B.~S., Linmans, J., Winkens, J., Cohen, T., and Welling, M.
\newblock Rotation equivariant cnns for digital pathology.
\newblock In \emph{International Conference on Medical image computing and
  computer-assisted intervention}, pp.\  210--218. Springer, 2018.

\bibitem[Verbeke \& Molenberghs(2005)Verbeke and Molenberghs]{book}
Verbeke, G. and Molenberghs, G.
\newblock \emph{Linear Mixed Models For Longitudinal Data}.
\newblock 01 2005.
\newblock ISBN 978-1-4419-0299-3.
\newblock \doi{10.1007/978-1-4419-0300-6}.

\bibitem[Wakefield(2013)]{bayesian_book}
Wakefield, J.
\newblock \emph{Bayesian and Frequentist Regression Methods}.
\newblock 01 2013.
\newblock ISBN 978-1-4419-0924-4.
\newblock \doi{10.1007/978-1-4419-0925-1}.

\bibitem[Ware(1985)]{longitud_cor_2}
Ware, J.~H.
\newblock Linear models for the analysis of longitudinal studies.
\newblock \emph{The American Statistician}, 39\penalty0 (2):\penalty0 95--101,
  1985.

\bibitem[Welling \& Teh(2011)Welling and Teh]{welling2011bayesian}
Welling, M. and Teh, Y.~W.
\newblock Bayesian learning via stochastic gradient {Langevin} dynamics.
\newblock In \emph{Proceedings of the 28th international conference on machine
  learning (ICML-11)}, pp.\  681--688, 2011.

\bibitem[Wenzel et~al.(2020)Wenzel, Roth, Veeling, Swiatkowski, Tran, Mandt,
  Snoek, Salimans, Jenatton, and Nowozin]{wenzel2020good}
Wenzel, F., Roth, K., Veeling, B., Swiatkowski, J., Tran, L., Mandt, S., Snoek,
  J., Salimans, T., Jenatton, R., and Nowozin, S.
\newblock How good is the bayes posterior in deep neural networks really?
\newblock In \emph{International Conference on Machine Learning}, pp.\
  10248--10259. PMLR, 2020.

\bibitem[Wilson(2020)]{wilson2020case}
Wilson, A.~G.
\newblock The case for bayesian deep learning.
\newblock \emph{arXiv preprint arXiv:2001.10995}, 2020.

\bibitem[Wu et~al.(2020)Wu, Zhang, Valiant, and R{\'e}]{Wu2020OnTG}
Wu, S., Zhang, H.~R., Valiant, G., and R{\'e}, C.
\newblock On the generalization effects of linear transformations in data
  augmentation.
\newblock In \emph{ICML}, 2020.

\bibitem[Yuan et~al.(2021)Yuan, Chen, Chen, Codella, Dai, Gao, Hu, Huang, Li,
  Li, Liu, Liu, Liu, Lu, Shi, Wang, Wang, Xiao, Xiao, Yang, Zeng, Zhou, and
  Zhang]{yuan2021florence}
Yuan, L., Chen, D., Chen, Y.-L., Codella, N., Dai, X., Gao, J., Hu, H., Huang,
  X., Li, B., Li, C., Liu, C., Liu, M., Liu, Z., Lu, Y., Shi, Y., Wang, L.,
  Wang, J., Xiao, B., Xiao, Z., Yang, J., Zeng, M., Zhou, L., and Zhang, P.
\newblock Florence: A new foundation model for computer vision, 2021.

\bibitem[Zeno et~al.(2020)Zeno, Golan, Pakman, and Soudry]{zeno2020cold}
Zeno, C., Golan, I., Pakman, A., and Soudry, D.
\newblock Why cold posteriors? on the suboptimal generalization of optimal
  bayes estimates.
\newblock \emph{3rd Symposium on Advances in Approximate Bayesian Inference},
  2020.

\bibitem[Zhai et~al.(2021)Zhai, Kolesnikov, Houlsby, and
  Beyer]{zhai2021scaling}
Zhai, X., Kolesnikov, A., Houlsby, N., and Beyer, L.
\newblock Scaling vision transformers, 2021.

\bibitem[Zhang et~al.(2020)Zhang, Li, Zhang, Chen, and
  Wilson]{zhang2019cyclicalsgmcmc}
Zhang, R., Li, C., Zhang, J., Chen, C., and Wilson, A.~G.
\newblock Cyclical stochastic gradient {MCMC} for {Bayesian} deep learning.
\newblock In \emph{International Conference on Learning Representations (ICLR
  2020)}, 2020.

\end{thebibliography}
\bibliographystyle{icml2022}

\newpage
\appendix
\onecolumn
\section{Omitted Proofs}
In this section we detail all the calculations and proofs omitted in the main text.
\subsection{Proof of Thm.~\ref{thm:correlated_means}}
\label{proof:Thm}
We have $n$ independent samples $x_1, \dots, x_n$ and we consider an additive augmentation $\tilde{x}^{b}_i=R_{\eta}(x_i) = x_i + \eta_i^{b}$ where $\eta_i^{b} \sim \mathcal{N}(0, \sigma_{\eta}^2)$ for $b=1,\dots,B$. Per datapoint $x_i$, we have $B$ augmentations $\tilde{x}_i^{1}, \dots, \tilde{x}_i^{B}$ originating from $x_i$ with independent underlying noise $\eta_i^{1}, \dots, \eta_i^{B}$. Augmentations of the same input display correlations which we can easily calculate
\begin{equation*}
    \begin{split}
        \operatorname{cov}(\tilde{x}_i^{b}, \tilde{x}_i^{b'}) &= \mathbb{E}[\tilde{x}_i^{b}\tilde{x}_i^{b'}] - \mathbb{E}[\tilde{x}_i^{b}]\mathbb{E}[\tilde{x}_j^{b'}] = \mathbb{E}[x_i^2] + \mathbb{E}[\eta_i^{b} x_i] + \mathbb{E}[\eta_i^{b'} x_i] + \mathbb{E}[\eta_i^{b} \eta_i^{b'}]-\mathbb{E}[x_i + \eta_i^{b}] \mathbb{E}[x_i + \eta_i^{b'}]\\
        &= \mu^2 + \sigma^2 + \mathds{1}_{\{b=b'\}}\sigma_{\eta}^2 - \mu^2\\
        &= \sigma^2 + \mathds{1}_{\{b=b'\}}\sigma_{\eta}^2 
    \end{split}
\end{equation*}
where we used the independence between $\eta_i^{b}, \eta_i^{b'}$ and $x_i$. On the other hand, augmentations from different samples remain independent:
\begin{equation*}
    \begin{split}
        \operatorname{cov}(\tilde{x}_i^{b}, \tilde{x}_j^{b'}) &= \mathbb{E}[\tilde{x}_i^{b}\tilde{x}_j^{b'}] - \mathbb{E}[\tilde{x}_i^{b}]\mathbb{E}[\tilde{x}_j^{b'}] = \mathbb{E}[{x}_i{x}_j] + \mathbb{E}[\eta_i^{b} x_j] + \mathbb{E}[\eta_j^{b'} x_i] + \mathbb{E}[\eta_i^{b} \eta_j^{b'}]-\mathbb{E}[x_i + \eta_i^{b}] \mathbb{E}[x_j + \eta_i^{b'}] \\
        &= \mu^2 - \mu^2 \\
        &= 0
    \end{split}
\end{equation*}
Collecting all $x_i^{b}$ into one vector $\bm{\tilde{x}} \in \mathbb{R}^{Bn}$, we can describe the joint distribution as
$$\bm{\tilde{x}}|\mu \sim \mathcal{N}(\mu \bm{1}_{Bn}, \bm{\tilde{\Sigma}})$$
where $\bm{\tilde{\Sigma}} \in \mathbb{R}^{Bn \times Bn}$ is block-diagonal with blocks $\bm{\Sigma} = \sigma_{\eta}^2 \mathds{1} + \sigma^2 \bm{1}_{Bn}\bm{1}_{Bn}^{T}$. We know by Bayes rule that the posterior is proportional to 
$$p(\mu|\bm{\tilde{x}}) \propto p(\bm{\tilde{x}}|\mu)p(\mu)$$
Instead of factorizing $p(\bm{\tilde{x}}|\mu)$ we just work with the joint distribution directly:
\begin{equation*}
\begin{split}
    p(\mu|\bm{\tilde{x}}; \bm{\tilde{\Sigma}}) &\propto e^{-\frac{1}{2}\left(\bm{\tilde{x}}-\mu\bm{1}_{Bn}\right)^{T}\bm{\tilde{\Sigma}}^{-1}\left(\bm{\tilde{x}}-\mu\bm{1}_{Bn}\right)}e^{-\frac{1}{2\sigma_0^2}(\mu-\mu_0)^2} 
    = e^{-\frac{1}{2}\mu^2\left(\bm{1}_{Bn}^{T}\bm{\tilde{\Sigma}}^{-1}\bm{1}_{Bn} + \frac{1}{\sigma_0^2}\right) + \mu\left(\bm{1}_{Bn}\bm{\tilde{\Sigma}}^{-1}\bm{x} + \frac{\mu_0}{\sigma_0^2}\right)}\\
    &\propto \mathcal{N}\left(\mu_{\text{post}}^{\bm{\tilde{\Sigma}}}, \left(\sigma_{\text{post}}^{\bm{\tilde{\Sigma}}}\right)^2\right)  
\end{split}
\end{equation*}
with the posterior mean and variance estimate
$$\mu_{\text{post}}^{\bm{\tilde{\Sigma}}} = \frac{\bm{1}_{Bn}^{T}\bm{\tilde{\Sigma}}^{-1}\bm{\tilde{x}} + \frac{\mu_0}{\sigma_0^2}}{\bm{1}_{Bn}^{T}\bm{\tilde{\Sigma}}^{-1}\bm{1}_{Bn} + \frac{1}{\sigma_0^2}} = \frac{\frac{1}{\sigma_{\eta}^2 + B\sigma^2}\sum_{i=1}^{Bn}\tilde{x}_i + \frac{\mu_0}{\sigma_0^2}}{\frac{Bn}{\sigma_{\eta}^2 + B\sigma^2} + \frac{1}{\sigma_0^2}}  \hspace{10mm} \left(\sigma_{\text{post}}^{\bm{\tilde{\Sigma}}}\right)^2 = \frac{1}{\bm{1}_{Bn}^{T}\bm{\tilde{\Sigma}}^{-1}\bm{1}_{Bn} + \frac{1}{\sigma_0^2}} = \frac{1}{\frac{Bn}{\sigma_{\eta}^2 + B\sigma^2} + \frac{1}{\sigma_0^2}}$$
We can perform the same calculation for tempered likelihoods assuming (wrongly) that the likelihood factorizes, leading to the tempered posterior
$$p_T(\mu|\bm{\tilde{x}};\mathds{1}) = \mathcal{N}(\mu_T, \sigma_T^2)$$
with tempered posterior statistics
$$\mu_T = \left(\frac{Bn}{T(\sigma^2 + \sigma_{\eta}^2)} + \frac{1}{\sigma_0^2}\right)^{-1}\left(\frac{\mu_0}{\sigma_0^2}+\frac{1}{T(\sigma^2 + \sigma_{\eta}^2)}\sum_{i=1}^{Bn}\tilde{x}_i\right) \hspace{10mm} \sigma_{T}^2 = \left(\frac{Bn}{T(\sigma^2 + \sigma_{\eta}^2)} + \frac{1}{\sigma_0^2}\right)^{-1}$$
We can check that setting the temperature as 
$$T = \frac{\sigma^2_{\eta}+ B\sigma^2}{\sigma^2_{\eta}+\sigma^2} $$
leads to an equality in distribution, i.e. $p_T(\mu|\bm{x};\mathds{1}) = p(\mu|\bm{\tilde{x}}; \bm{\tilde{\Sigma}}) $
\subsection{Correlations in Regression Setting}
\label{app:correlations}
We can calculate the correlation in the regression setting as follows:
\begin{equation*}
\begin{split}
    \operatorname{cov}(\tilde{\epsilon}_i, \tilde{\epsilon}_j) &= \mathbb{E}[\tilde{\epsilon}_i\tilde{\epsilon}_j] = \mathbb{E}[(\epsilon + \delta_{\bm{\eta}_i})(\epsilon + \delta_{\bm{\eta}_j})] = \mathbb{E}[\epsilon^2] = \sigma^2_{\epsilon}
\end{split}
\end{equation*}
where we used that $\tilde{\epsilon}$'s are centered by assumption since $\delta$'s are centered. On the other hand, we can compute the respective variance as 
\begin{equation*}
    \operatorname{var}(\tilde{\epsilon}_i) = \mathbb{E}[(\epsilon + \delta_{\bm{\eta}_i})^2] = \mathbb{E}[\epsilon^2] + \mathbb{E}[\delta_{\bm{\eta}_i}^2] = \sigma^2 + \operatorname{var}(\delta_{\bm{\eta}})
\end{equation*}
The resulting correlation is thus given by 
$$\operatorname{cor}(\tilde{\epsilon}_i, \tilde{\epsilon}_j) = \frac{\operatorname{cov}(\tilde{\epsilon}_i, \tilde{\epsilon}_j)}{\operatorname{var}(\tilde{\epsilon}_i)} = \frac{\sigma^2_{\epsilon}}{\sigma^2_{\epsilon} + \operatorname{var}(\delta_{\bm{\eta}})}$$
\subsection{Correlations in Logistic Regression}
\label{app:logistic}
Correlations are not an artifact of the regression setting but also arise in classification tasks. To that end, we rely on the latent variable model of logistic regression, where a latent variable 
$$z = f_{\bm{\theta}_{*}}(\bm{x}) + \epsilon$$ 
is introduced, with $\epsilon \sim \operatorname{Logistic}(0, 1)$ following a centered logistic variable with variance. These latent variables $z$ then give rise to the response through the relation 
$$y=\mathds{1}_{\{z\geq0\}}.$$ 
In perfect duality to the regression setting, the errors implied by augmentations of the same covariate $\bm{x}$ exhibit significant correlations. More concretely, we again augment as $\bm{\tilde{x}} = R_{\bm{\eta}}(\bm{x})$ and set $\tilde{y} = y$. Here we have a latent variable $\tilde{z}$ which might differ from $z$. Notice that in the case of $\tilde{z}=z$, we are in the same setting as regression. In general, this might however not be true but we know that for two augmentations $\bm{\eta}_i$ and $\bm{\eta}_j$ with errors $\tilde{\epsilon}_i$ and $\tilde{\epsilon}_j$, the following relation holds:
$$\tilde{y}_i = y = \tilde{y}_j \iff \mathds{1}_{\{f_{\bm{\theta}_{*}}(R_{\bm{\eta}_i}(\bm{x})) + \tilde{\epsilon}_i \geq 0\}} = \mathds{1}_{\{f_{\bm{\theta}_{*}}(\bm{x}) + \epsilon \geq 0\}} = \mathds{1}_{\{f_{\bm{\theta}_{*}}(R_{\bm{\eta}_j}(\bm{x})) + \tilde{\epsilon}_j \geq 0\}} \hspace{3mm} \text{a.s.}$$
Hence, the two errors are related through the equation
\begin{equation*}
    \mathds{1}_{\{f_{\bm{\theta}_{*}}(R_{\bm{\eta}_i}(\bm{x})) + \tilde{\epsilon}_i \geq 0\}} = \mathds{1}_{\{f_{\bm{\theta}_{*}}(R_{\bm{\eta}_j}(\bm{x})) + \tilde{\epsilon}_j \geq 0\}} \hspace{3mm} \text{a.s.}
\end{equation*}
and hence $\tilde{\epsilon}_i$ and $\tilde{\epsilon}_j$ share significant correlation. In case of perfect invariance, i.e.$f_{\bm{\theta}_{*}}(R_{\bm{\eta}_i}(\bm{x})) = f_{\bm{\theta}_{*}}(R_{\bm{\eta}_j}(\bm{x})) = f_{\bm{\theta}_{*}}(\bm{x})$, we find that 
\begin{equation*}
    \mathds{1}_{\{ \tilde{\epsilon}_i \geq -f_{\bm{\theta}_{*}}(\bm{x})\}} = \mathds{1}_{\{ \tilde{\epsilon}_j \geq -f_{\bm{\theta}_{*}}(\bm{x})\}} \hspace{3mm} \text{a.s.}
\end{equation*}
which, for infinitely supported random variables such as the logistic distribution, can only hold if $\tilde{\epsilon}_i = \tilde{\epsilon}_j$ a.s., hence leading to perfectly correlated errors.
\subsection{Correlation in Linear Regression}
\label{app:linear_correlation}
As seen in the main text, it holds that  
$$\tilde{\epsilon} = \epsilon + \delta_{\bm{\eta}}$$
where $\delta_{\bm{\eta}} = -\bm{\eta}^{T}\bm{x}$. Since $\bm{\eta}\sim \mathcal{N}(\bm{0}, \bm{\Sigma}_{\eta})$, it is evident that $\mathbb{E}[\delta_{\bm{\eta}}] = 0$ since $\mathbb{E}[\bm{\eta}] = \bm{0}$. Moreover
$$\operatorname{var}(\tilde{\epsilon}) = \sigma_{\epsilon}^2 + \mathbb{E}[\bm{x}^{T}\bm{\eta}\bm{\eta}^{T}\bm{x}] = \sigma_{\epsilon}^2 + \bm{x}^{T}\bm{\Sigma}_{\eta}\bm{x}$$
Finally, for two augmentations $\bm{\eta}_i$, $\bm{\eta}_j$, it  holds for the respective errors that $\tilde{\epsilon}_i,\tilde{\epsilon}_j$ that
\begin{equation*}
    \begin{split}
        \operatorname{cov}(\tilde{\epsilon}_i,\tilde{\epsilon}_j) = \mathbb{E}[\tilde{\epsilon}_i\tilde{\epsilon}_j] = \mathbb{E}[\epsilon^2] = \sigma^2_{\epsilon}
    \end{split}
\end{equation*}
We can hence conclude that the correlation is given by 
$$\operatorname{cor}(\tilde{\epsilon}_i,\tilde{\epsilon}_j) = \frac{\sigma^2_{\epsilon}}{\sigma_{\epsilon}^2 + \bm{x}^{T}\bm{\Sigma}_{\eta}\bm{x}}$$

\subsection{Intermediate Values for Correlation}
\label{app:inter-correlations}
In this section we study to what degree tempering with $T$ can reduce the misspecification stemming from ignoring the correlation between errors. The posteriors in both the i.i.d. and correlation setting are intractable for general models $f_{\bm{\theta}}$ but the two likelihoods take a simple form. We show that in general, no temperature $T$ can be found to make the two likelihoods match, i.e. their KL-divergence is not $0$. This in turn implies that the posteriors cannot match either (both models use the same prior). \\
For simplicity we again assume a linear regression setting with one sample $(\bm{x}, y)$ and additive augmentations, i.e. $f_{\bm{\theta}}(\bm{x}) = \bm{\theta}^{T}\bm{x}$ and $G_{\bm{\eta}}(\bm{x}) = \bm{x}+\bm{\eta}$ for $\bm{\eta} \sim \mathcal{N}(\bm{0}, \sigma_{\eta}^2\mathds{1})$. As a result, we find that the augmentation errors $\tilde{\epsilon}_1, \dots, \tilde{\epsilon}_B$ still follow a joint Gaussian distribution with
$$\bm{\tilde{\epsilon}} \sim \mathcal{N}(\bm{0}, \bm{\tilde{\Sigma}})$$
with $\tilde{\Sigma}_{ii} = \sigma_{\epsilon}^2 + \sigma_{\eta}^2$ and $\tilde{\Sigma}_{ij} = \sigma_{\epsilon}^2$. On the other hand, we can also decide to ignore the correlations and temper, i.e. just work with $\frac{\sigma_{\epsilon}^2 + \sigma_{\eta}^2}{T}\mathds{1}$, and thus factorize the likelihood. In both cases, we have a Gaussian likelihood of the form
$$p_T(\bm{y}|\bm{x}, \bm{\theta}; \mathds{1}) \sim \mathcal{N}(\bm{\tilde{X}}\bm{\theta}, (\sigma_{\epsilon}^2 + \sigma_{\eta}^2)\mathds{1}) \hspace{5mm} p(\bm{y}|\bm{x}, \bm{\theta}; \bm{\tilde{\Sigma}}), \sim \mathcal{N}(\bm{\tilde{X}}\bm{\theta}, \bm{\tilde{\Sigma}}))$$
where we define the matrix of augmentations $\bm{\tilde{X}} \in \mathbb{R}^{B \times d}$. \\[2mm] In general, we can express the KL-divergence between two Gaussians $\rho_1 \sim \mathcal{N}(\bm{\mu}_1, \bm{\Sigma}_1)$ and $\rho_2 \sim \mathcal{N}(\bm{\mu}_2, \bm{\Sigma}_2)$ as
\begin{equation}
    \operatorname{KL}(\rho_1 \hspace{1mm}|| \hspace{1mm}\rho_2) = \frac{1}{2}\left[\log \frac{|\bm{\Sigma}_2|}{|\bm{\Sigma}_1|} -B + Tr(\bm{\Sigma}_2^{-1}\bm{\Sigma}_1) + (\bm{\mu}_2 - \bm{\mu}_1)^T\bm{\Sigma}_2^{-1}(\bm{\mu}_2 - \bm{\mu}_1)\right] .
\end{equation}
In our case, let $\rho_1 \sim \mathcal{N}(\bm{\tilde{X}}\bm{\theta}, \frac{\sigma^2_{\epsilon} + \sigma_{{\eta}}^2}{T}\mathds{1})$ be the tempered likelihood and $\rho_2 \sim \mathcal{N}(\bm{\tilde{X}}\bm{\theta}, \bm{\tilde{\Sigma}})$ the correlated likelihood. 
We have that $|\frac{\sigma^2_{\epsilon} + \sigma_{{\eta}}^2}{T}\mathds{1}| = \left(\frac{\sigma^2_{\epsilon} + \sigma_{{\eta}}^2}{T}\right)^B$ and therefore the derivative of its logarithm w.r.t $T$ is:
\begin{equation}
    \frac{\partial}{\partial T} B \log \frac{\sigma^2_{\epsilon} + \sigma_{{\eta}}^2}{T} = - \frac{B}{T} .
\end{equation}

For the trace term we have that:
\begin{equation}
    \frac{\partial}{\partial T} Tr\left(\frac{(\sigma^2_{\epsilon} + \sigma_{{\eta}}^2)\bm{\tilde{\Sigma}}^{-1}}{T}\right) = - \frac{\sigma^2_{\epsilon} + \sigma_{{\eta}}^2}{T^2} Tr(\bm{\tilde{\Sigma}}^{-1})
\end{equation}

So, by setting the derivative of the KL to zero and solving for $T$:

\begin{equation}
    T^{*} = (\sigma^2_{\epsilon} + \sigma_{{\eta}}^2)\frac{Tr(\bm{\tilde{\Sigma}}^{-1})}{B}
\end{equation}
Plugging-in the optimal value $T^{*}$ into the KL-divergence gives
$$2\operatorname{KL}\left(p_{T^{*}}(\bm{y}|\bm{x}, \bm{\theta}; \mathds{1}) \hspace{1mm}|| \hspace{1mm}p(\bm{y}|\bm{x}, \bm{\theta}; \bm{\tilde{\Sigma}})\right) = \log(|\bm{\tilde{\Sigma}}|) - B \log\left(\frac{B}{\operatorname{Tr}\left(\bm{\tilde{\Sigma}}^{-1}\right)}\right) \stackrel{!}{=}0$$
or equivalently 
$$\log(|\bm{\tilde{\Sigma}}|) \stackrel{!}{=}B\log\left(\frac{B}{\operatorname{Tr}\left(\bm{\tilde{\Sigma}}^{-1}\right)}\right) \iff |\bm{\tilde{\Sigma}}| \stackrel{!}{=} \left(\frac{B}{\operatorname{Tr}\left(\tilde{\bm{\Sigma}}^{-1}\right)}\right)^{B}$$
We can plug-in the concrete $\bm{\tilde{\Sigma}} = \sigma_{\eta}^2 \mathds{1} + \sigma_{\epsilon}^2\bm{1}_B\bm{1}_B^{T}$ to find
$$(\sigma_{\eta}^2 + B \sigma_{\epsilon}^2) \sigma_{\eta}^{2(B-1)} \stackrel{!}{=} \left(\frac{\sigma_{\eta}^2(\sigma_{\eta}^2 + \sigma_{\epsilon}^2B)}{\sigma_{\eta}^2 + \sigma_{\epsilon}^2(B-1)}\right)^{B}$$
We observe that if $B>1$ this equation cannot hoald for a general $\sigma_{\eta}^2$. On the other hand, $\sigma_{\eta}^2$6 governs the correlation coefficient , which was shown in Appendix~\ref{app:linear_correlation} to be 
$$\operatorname{cor}(\tilde{\epsilon}_i, \tilde{\epsilon}_j) = \frac{\sigma_{\epsilon}^2}{\sigma_{\epsilon}^2 + \sigma_{\eta}^2}$$
Tempering can hence not match the likelihoods exactly for a general correlation (i.e. general $\sigma_{\eta}^2$) and hence the posteriors cannot match either.
 $1$ and $B$ !

\section{Experimental Details}
\label{app:experiments}
In this section we describe the experimental setup, including the architectural details and the SG-MCMC hyperparameters. For the SG-MCMC sampler, we adapted the code from \citet{wenzel2020good}\footnote{\url{https://github.com/google-research/google-research/tree/master/cold_posterior_bnn}}. For the implementation of the group equivariant layers, we used the code from \citet{veeling2018rotation}\footnote{\url{https://github.com/basveeling/keras-gcnn}}

\subsection{SG-MCMC}
For the experiments with ResNet20 and G-ResNets on CIFAR-10, we have the following hyperparameters:
\begin{itemize}
    \item initial learning rate: $0.1$
    \item burn-in period: $150$ epochs
    \item cycle length: $50$ epochs
    \item total training time: $1500$ epochs
\end{itemize}

For the experiments with convolutional and G-convolutional networks on CIFAR-10, we have the following hyperparameters:
\begin{itemize}
    \item initial learning rate: $0.5$
    \item burn-in period: $150$ epochs
    \item cycle length: $50$ epochs
    \item total running time: $1000$ epochs
\end{itemize}

Note that for the experiments where only the likelihood is tempered, the reparamterization of the SG-MCMC updates explained in Section \ref{sec:exps} force one to adapt the learning rate to the temperature: if we increase the temperature by 10 times, we should make the learning 10 times larger. Therefore the learning rate specified above refers to the learning rate at $T=1$. We stress that we adapt the learning rate for a fair comparison of runs with different temperature, but changing the learning rate does \emph{not} affect the posterior distribution, while the temperature does. We refer to \citet{aitchison2020statistical} for the details of a similar argument in the different context of dataset curation. 

Finally, the experiments are executed on Nvidia DGX-1 GPU nodes equipped with 4 20-core Xeon E5-2698v4 processors, 512 GB of memory and 8 Nvidia V100 GPUs. 

\subsection{Neural Network Architectures}
For the SG-MCMC experiments, we use a 20-layer architecture with residual layers \citep{resnet2016} and batch normalization. The architecture with only convolutional layers is composed of 4 convolutional layers, the first two with 32 filters and the last two with 64 filters. We experiments both with 2-strided and 1-strided convolutions, as detailed in Section \ref{sec:exps}. The batch size is 128 across all experiments. 
The group equivariant architectures (G-ResNet and G-Conv) are designed by replacing the convolutional layers and pooling layer with the Group equivariant counterpart, as detailed in \cite{cohen2016group}. To build more invariant architectures, also G-batch normalization layers should replace batch norm. However, due to a code incompatibility between the two code repositories (SG-MCMC and Group equivariant layers) we are not able to use G-batch norms. 

\subsection{Data Augmentation}
The datapoints are always transformed in an online fashion and not precomputed, as it is commonly done in deep learning for memory efficiency. This means that when a batch of original datapoints $S_t$ is processed at epoch $t$, then each datapoint is preprocessed with the augmentation function and then propagated through the network. This means that at every epoch we potentially have new augmented datapoints. Therefore the total the number of augmentationd per datapoint is equal to the number of epochs. On the contrary, for the experiment with fixed number of augmentations, we fix a sequence of $B$ random seeds, which is then repeated every $B$ epochs, guaranteeing that the total number of augmentations is fixed.

\section{Further Results}
Here we show the figures of the plot for all the experiments of Section \ref{sec:exps}. 

\begin{figure}[h!]
\centering
\includegraphics[width=0.45\textwidth]{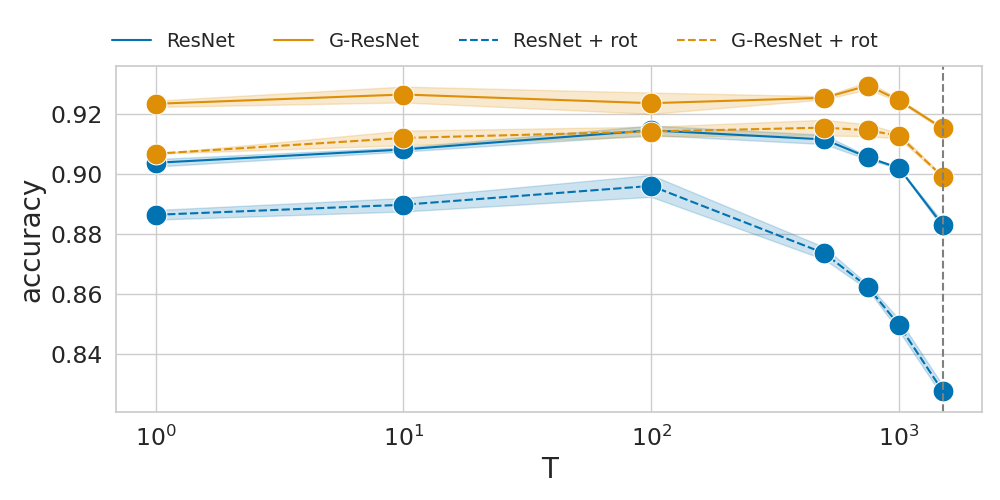}
\caption{Test accuracy as a function of the temperature $T$ for a ResNet with standard convolutions (blue) and G-convolutions (orange), with or without extra 90$\degree$  rotations (dashed lines).}
\label{fig:acc_gconv_vs_conv}
\end{figure}

\paragraph{Hot or Cold Posteriors?}
In Fig. \ref{fig:acc_gconv_vs_conv}, we plot the accuracy results for \textit{ResNet20} for the tempered posterior and tempered likelihood cases.

\begin{figure}[h!]
\centering
\includegraphics[width=0.45\textwidth]{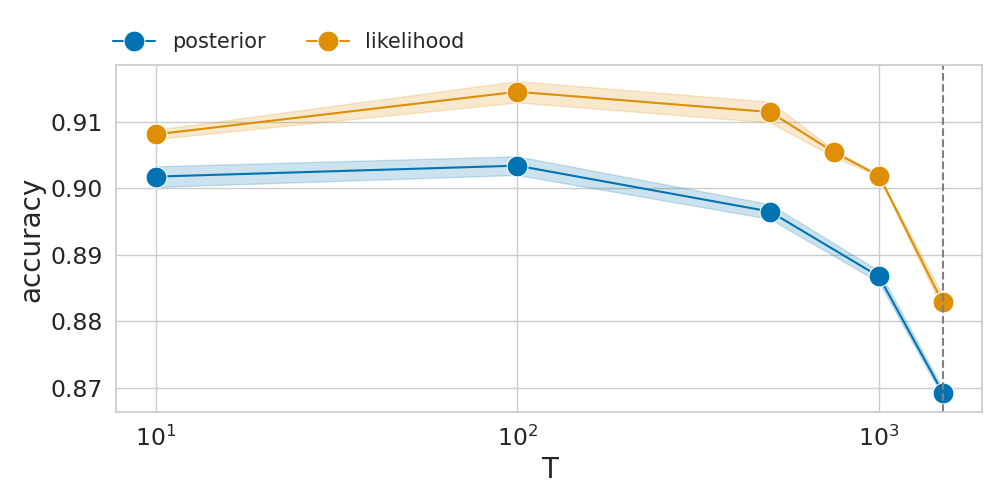}
\caption{Test accuracy as a function of the temperature $T$ for posterior tempering (blue) and likelihood tempering (orange).} 
\label{fig:acc_lik_vs_posterior_temp}
\end{figure}

\begin{figure}[h!]
\centering
\includegraphics[width=0.45\textwidth]{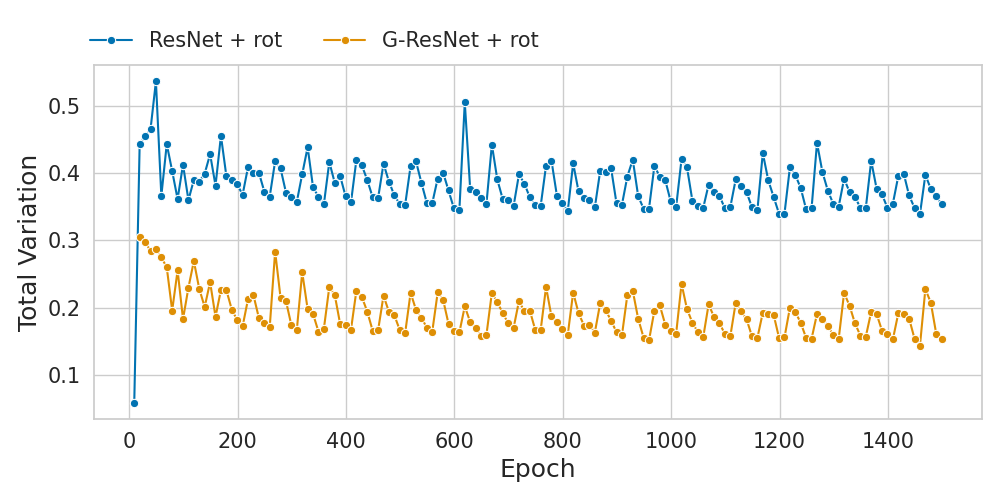}
\caption{Total variation during learning for the experiments with G-ResNet20. A lower value of the total variation indicates a greater degree of invariance.}
\label{fig:inv_plot_resnet}
\end{figure}

\paragraph{G-Convolutions alleviates the CPE}
In Fig. \ref{fig:acc_gconv_vs_conv}, we plot the accuracy results for \textit{ResNet20} and \textit{G-ResNet20} trained with and without additional 90$\degree$ rotations. Additionally, in Fig. \ref{fig:inv_plot_resnet}, we plot the evolution of the invariance measure (total variation) for the \textit{G-ResNet} trained with and without additional 90$\degree$ rotation.

\begin{figure}[h!]
\centering
\includegraphics[width=0.45\textwidth]{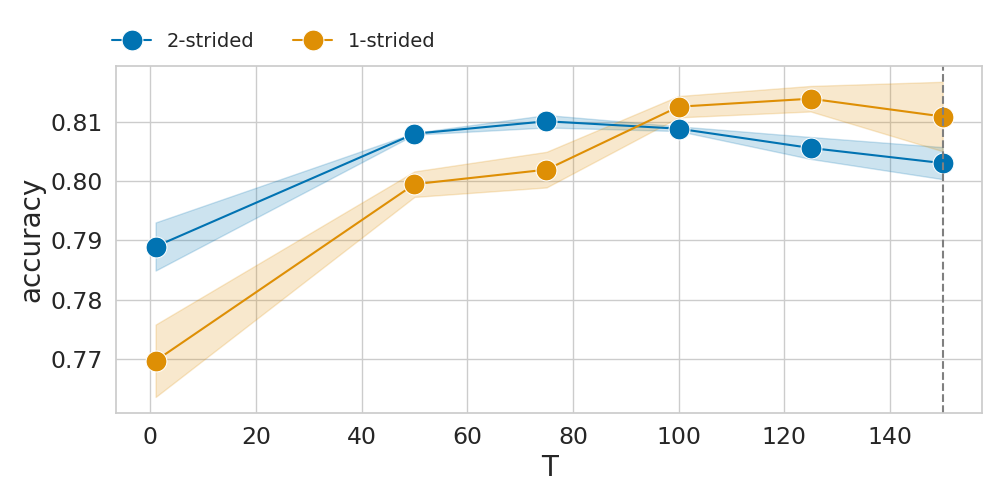}
\caption{Test accuracy as a function of the temperature for a model trained 2-strided (blue) and 1-strided (orange) G-convolutional network.  }
\label{fig:acc_gconv_inv}
\end{figure}

\begin{figure}[h!]
\centering
\includegraphics[width=0.45\textwidth]{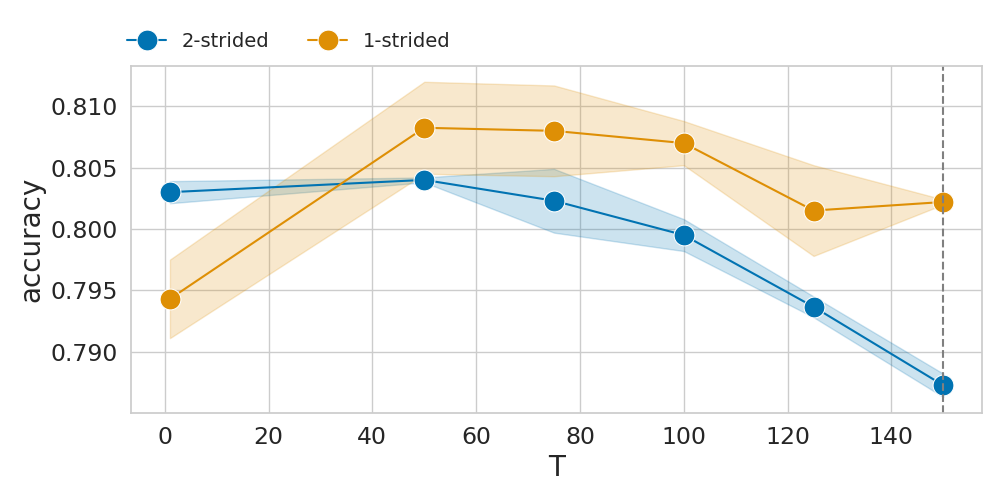}
\caption{Same setting as in Fig. \ref{fig:acc_gconv_inv}, but this time we perform an extra random rotation of 10$\degree$. Note how the loss of invariance shifts the optimal temperature significantly, in both the 2-strided and 1-strided case.}
\label{fig:acc_gconv_small_rot}
\end{figure}

\paragraph{An Almost Invariant Model}
Finally, in Fig. \ref{fig:acc_gconv_inv} and Fig. \ref{fig:acc_gconv_small_rot} we plot the accuracy for the G-Convolutional networks trained with flip and 90$\degree$ rotations and with additional random 10$\degree$ rotations. 

\end{document}